\pdfoutput=1

\documentclass[11pt]{article} 

\usepackage{acl}
\usepackage{float}
\usepackage{latexsym}
\usepackage{listings}
\usepackage[T1]{fontenc}
\usepackage{microtype,cuted,longtable}
\usepackage{times,multicol}
\usepackage{url,tcolorbox}
\usepackage{paralist,multicol,xtab}
\usepackage{graphicx}
\usepackage{wrapfig}
\usepackage{fancyvrb}
\usepackage[inline]{enumitem}
\usepackage{amssymb,multirow}
\usepackage{natbib}
\usepackage{booktabs,fvextra }
\usepackage{comment}
\usepackage{subcaption}
\usepackage{amssymb}
\usepackage{verbatim}
\usepackage{longtable}
\usepackage{tikz}
\usepackage{glossaries}
\usepackage{array}
\usepackage{enumitem,verbatimbox}
\usepackage{inconsolata}
\usepackage[a4paper,margin=2.5cm]{geometry}
\usepackage{xspace}
\usepackage{algorithm} 
\usepackage{algpseudocode}

\newcommand{\fig}{Fig.\xspace}
\newcommand{\tab}{Tab.\xspace}
\newcommand{\critic}{$C=\{c_1,\dots, c_n\}$\xspace}
\newcommand{\criterion}{$c_i: \{\omega_j\}_{j=1}^m$}
\newcommand{\utility}{$U_t(s)=\{Q_i(s|c_i)\}_{i=1}^n$\xspace}
\newcommand{\qa}{\emph{QuantifierAgent}\xspace}
\newcommand{\ca}{\emph{CriticAgent}\xspace}
\newcommand{\va}{\emph{VerifierAgent}\xspace}

\lstdefinelanguage{json-like}{
  basicstyle=\ttfamily,
  numbers=left,
  numberstyle=\tiny\color{gray},
  numbersep=5pt,
  showstringspaces=false,
  breaklines=true,
  frame=single,
  morestring=[b]",
  stringstyle=\color{blue},
  morecomment=[l]{//},
  commentstyle=\color{green},
  morecomment=[s]{/*}{*/},
  morekeywords={content, role}
}

\newcounter{todocnt}

\title{Assessing and Verifying Task Utility in LLM-Powered Applications}

\author{
\textbf{Negar Arabzadeh}$^{1}$\thanks{\phantom{\_\_}Work done during an internship at Microsoft Research} \quad 
\textbf{Siqing Huo}$^1$ \quad 
\textbf{Nikhil Mehta}$^3$ \quad 
\textbf{Qingyun Wu}$^3$ \quad 
\textbf{Chi Wang}$^2$ \quad \\
\textbf{Ahmed Awadallah}$^2$ \quad  
\textbf{Charles L. A. Clarke$^1$} \quad
\textbf{Julia Kiseleva}$^2$  
\\
$^1$Univerity of Waterloo\\
$^2$Microsoft Research\\
$^3$Pennsylvania State University\\
$^4$Purdue University \\
}

\begin{document}

\maketitle
\begin{abstract}

The rapid development of Large Language Models (LLMs) has led to a surge in applications that facilitate collaboration among multiple agents, assisting humans in their daily tasks. However, a significant gap remains in assessing to what extent LLM-powered applications genuinely enhance user experience and task execution efficiency. This highlights the need to verify utility of LLM-powered applications, particularly by ensuring alignment between the application's functionality and end-user needs.
We introduce AgentEval, a novel framework designed to simplify the utility verification process by automatically proposing a set of criteria tailored to the unique purpose of any given application. This allows for a comprehensive assessment, \textit{quantifying the utility of an application against the suggested criteria}. We present a comprehensive analysis of the effectiveness and robustness of AgentEval for two open source datasets including Math Problem solving and ALFWorld House-hold related tasks. For reproducibility purposes, we make the data, code and all the logs publicly available at \url{https://bit.ly/3w3yKcS}

\end{abstract}

\section{Introduction}
\label{sec:intro}

\begin{figure}[!tb]
\centering
\vspace{-1em}
  \includegraphics[clip, trim=9.7cm 6.2cm 5cm 2.9cm,scale=0.61]{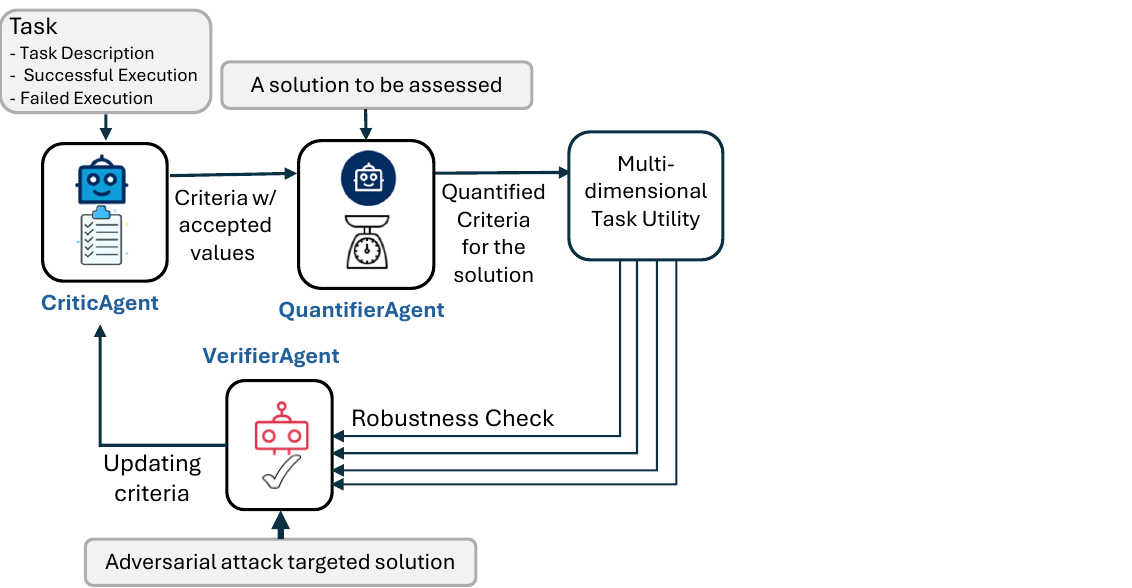}
\caption{An overview of the \emph{AgentEval} framework: \ca creates a set of criteria  and suggested values; \qa quantifies the criteria for a considered application; and \va verifies the criteria based on its robustness. The output of the \qa is a multi-dimensional assessment of the utility of the application based on a suggested list of criteria and their evaluations.}
\vspace{-1em}
\label{fig:agent-eval}
\end{figure}

One of the long-lasting goals for intelligent agents~\cite{winograd1972understanding} is for them to seamlessly interact with humans in natural language and help their end-users with their tasks, such as completing household tasks, math tutoring, and so on.
The rapid development of open-source libraries~\cite{wu2023autogen,li2023camel} helps that goal by simplifying the development of LLM-powered agentic applications for various user-centered tasks~\cite{liang2023encouraging,hong2023metagpt,talebirad2023multi}. To ensure that the application's behavior meets the requirements of the application developers, it is also crucial to assess its potential utility to end users~\cite{dibia2023aligning}, as this can significantly impact its improvement journey. Taking into account a range of applications, it is unrealistic to assume benchmarking for every domain, including but not limited to code generation~\cite{liu2024exploring}, health care~\cite{andrew2024potential}, and many others whose development we witness every day~\cite{wu2023autogen}. 
%
Moreover, directly evaluating agentic applications poses challenges, as current approaches predominantly rely on end-to-end success metrics i.e., whether the application accomplishes tasks~\cite{shridhar2020alfworld, DBLP:journals/corr/abs-1912-01734,pmlr-v229-myers23a}. However, understanding a user's interactions with an application involves much more than success alone~\cite{pmlr-v176-kiseleva22a, pmlr-v220-kiseleva22a, zhang2023creative}. Consider math problem solving, although it is important that the application solves the problem correctly, its ability to present and explain solutions based on various \emph{criteria}, such as completeness, conciseness, and clarity, is crucial. 
%
Furthermore, success is not always clearly defined for a task. Recognizing such criteria and being able to quantify them is essential to assess whether developer requirements are being satisfied and if the application brings utility to the end-users. Given the objective of assessing arbitrary applications, relying solely on end-to-end success metrics is untenable, due to the expansive range of tasks requiring automation. The question is \emph{how to design a flexible methodology to assess the task utility for diverse set of applications?}


To bridge this gap, we introduce \emph{AgentEval}, a framework to gauge the utility of LLM-powered applications. Its goal is to assess the utility by providing application developers with insights into how the current flow can be characterized. \emph{AgentEval} builds on recent work showing that LLMs can be a scalable and cost-effective alternative to human evaluation for open-ended tasks~\cite{li2023collaborative}.
\emph{AgentEval} as illustrated in \fig~\ref{fig:agent-eval}, consists of the three following agents, formally defined in Sec.~\ref{sec:agenteval}:
(1)~\ca suggests the list of criteria based on the task description and a pair of solutions, where one is preferred over the other one (e.g., successful and failed examples). For instance, for math problems, the criteria could be be {Efficiency} and {Clarity} of the proposed solution; 
(2)~\qa quantifies how the solution performs for each criterion and returns the utility function, e.g. for math problems, if the ' Clarity is `not clear', `moderately clear', or `very clear'; 
(3)~\va verifies the quality of the assessment of the suggested criteria to make sure the criteria are essential, robust, informative and have high discriminative power. 


In summary, our main contributions are:
\begin{enumerate}[leftmargin=*, nosep, label=\textbf{C\arabic*}]

\item Introducing \emph{AgentEval}, a novel framework that leverages LLM-powered agents as a scalable and cost-effective alternative to human evaluations, to produce task utility through the collaboration of three agents: \ca, \qa and \va; and
\item An in-depth analysis of \emph{AgentEval} robustness for two applications across different solutions, that can be replicated on an unseen domain.
\end{enumerate}

\noindent

\section{Related Work}
\label{sec:rel-work}


\subsection{Evaluation of LLMs}
\label{sec:llm-eval}
Prior work~\cite{guo2023evaluating, ziyu-etal-2023-lens, chang2023survey, liang2023holistic} has extensively studied the evaluation of LLMs on various fronts: how ethically sound they are \cite{stahl2024ethics}, how they align to human preferences~\cite{hendrycks2020aligning, kopf2024openassistant}, their  robustness~\cite{wang2023robustness}, and the knowledge, and reasoning capabilities they posses~\cite{bian2023chatgpt}.
Recent work evaluates LLMs on  more specialized tasks, such as medical domain \cite{jin2019pubmedqa}, multi-modal tasks~\cite{mialon2023gaia, bang2023multitask}, or as agents in interactive environments~\cite{liu2023agentbench}. 
 

\subsection{User satisfaction prediction}
\label{sec:sat-pred}
Studies suggest that users interacting with various systems operate with specific utility functions in mind~\cite{li2020optimizing, azzopardi2018measuring, ahmadvand2022supporting}. Traditionally, metrics defining user satisfaction were designed using large-scale collected behavioral signals~\cite{kiseleva2014modelling}, and were tailored to specific applications, such as intelligent assistants~\cite{kiseleva2016predicting, kiseleva2016understanding}, web search engines~\cite{ williams2016detecting,williams2016your,williams2017does}, dialogue systems~\cite{see2019makes}, multi-turn conversations~\cite{li2021deus} and general-purpose personal assistants~\cite{kiseleva2017evaluating}. 
It was demonstrated that assessing users' satisfaction requires goes beyond a single metric. As such, here, we propose a flexible framework to assess user and developer requirements, which can eventually be used to improve the application flow.

\subsection{Using LLMs as evaluators}
\label{sec:llm-evaluators}
More recently, there has been a growing trend in utilizing LLMs as evaluators~\cite{chiang2023can,fu2023gptscore}, such as for qualitative research~\cite{bano2023exploring}, or summarization.
Specifically, \citet{jain2023multi} studied the efficacy of few-shot prompted LLM evaluators in evaluating summaries that were written by other LLMs. Similarly, \citet{wang2023chatgpt} explore if ChatGPT itself can be used as an evaluator, by prompting it to score texts. Other works~\cite{tjuatja2023llms, liu2023voices, chiang2023can} look at how LLMs can be used as proxies for human behavior, or work with humans, such as CoEval~\cite{li2023collaborative}, which showed how LLMs can make human evaluation easier. \citet{pan2024autonomous} also show how LLM evaluators can help build models that increase performance on downstream task. Building on the above, a different line of works identify weaknesses in single LLMs as direct evaluators \cite{huang2023large}, and propose to improve them, such as a multi-step calibration framework~\cite{wang2023large}. 
Given these drawbacks, recent work has looked at how multiple LLM agents can be used as evaluators. \citet{chan2023chateval}, proposed ChatEval, a multi-agent team that discusses and evaluates responses from agents on generation tasks (debate-style), leading to text that aligns with better human preferences. Similarly, \citet{chern2024can} proposed a multiple agent-debate-assisted meta-evaluation framework.

\noindent
Building on these works, we propose an automatic multi-agent assessment of utility for arbitrary LLM-powered applications, to provide deep insights for developers. Our framework can uncover current flaws in these applications, and may lead to improvements in them, particularly if the application flow changes after it is applied, and then it is re-used.


\section{Task Utility}
\label{sec:agenteval}

\begin{figure}[!tb]
\centerline{%
\vspace{-1em}
  \includegraphics[clip, trim=6.1cm 3.5cm 11.2cm 4.4cm,scale=0.5]{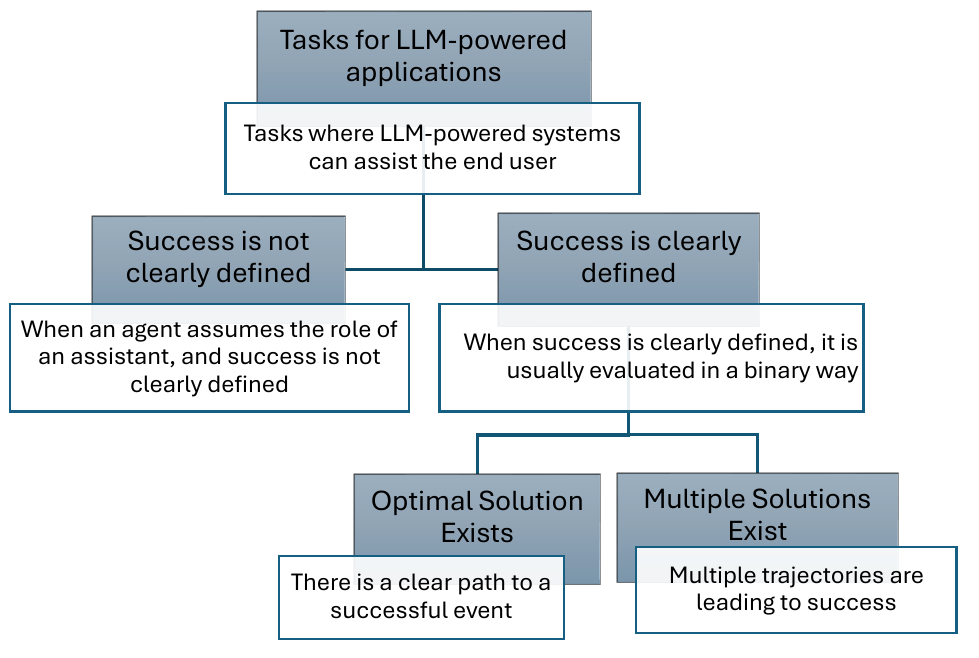}}
\caption{The taxonomy of tasks assessment.}
\vspace{-1em}
\label{fig:task-taxonomy}
\end{figure}

\fig~\ref{fig:task-taxonomy} outlines a taxonomy of target tasks for LLM-powered applications, in terms of success metrics. At a high level, these tasks can be categorized into:

\noindent \emph{1) Success is not clearly defined} --- Users use the system in an assistive manner, seeking suggestions from it, rather than expecting it to solve the task. For example, a user can request the system to generate an email. The user usually uses the system's response as a template, which can later be edited.
Directly evaluating assistive tasks like these is hard, particularly for online evaluation, or when dealing with less well-defined tasks. One potential approach is to directly ask users how useful the help was, but this is not well-calibrated~\cite{borisov2018calibration}, hard to quantify~\cite{sepliarskaia2018preference}, and expensive.

\noindent \emph{2) Success is clearly defined} --- It is clear whether the system solved the task or not, for example, assisting with household tasks, where success is clear and measurable. This category can be further divided into two subcategories:
\begin{itemize} [leftmargin=*, nosep]
\item \emph{an optimal solution exists} --- only one successful outcome is possible. For example, when asking an assistant to turn on a light, success is clearly defined, as there is only one way to do it.
\item \emph{multiple solutions exist} --- Increasingly, we observe situations where multiple trajectories of agent behavior can lead to success. For example, when asking an agent to suggest a food recipe, success could be multiple cuisines tasting good, but perhaps the recipe should not be expensive.
\end{itemize}

\noindent
\emph{AgentEval} is currently focused on tasks where \textit{success is clearly defined and multiple successful solutions} may exist. 

Previous research on assistive agents suggests human pairwise preferences as one of the most optimal assessments, i.e. when the annotator is presented with two agents side by side and asked for their preferences~\cite{pmlr-v220-kiseleva22a}. In this setup of side-by-side pairwise comparison, humans tend to suggest a list criteria, explaining why they prefer one agent over the other. For instance,`the first agent was faster' or `the second agent converses more naturally'. This comparative setup can guide humans to come up with a list of criteria that helps to infer the utility of the task. With this in mind, we designed \emph{AgentEval} (Fig.~\ref{fig:agent-eval}),
by employing LLMs to help us understand, verify, and assess task utility, namely:

\begin{itemize} [leftmargin=*, nosep]
\item \ca: The goal of this agent is to suggest a set of criteria that can be used to assess task utility. The \ca is given a task description, as well as optionally several pairs of solutions, where preferably some are preferred over the other ones, for instance, successful and failed examples. \ca would return a set of criteria \critic, where each criterion $c_i$ is accompanied by a set of accepted values $\omega$ as \criterion. For example, for solving math problems, the \ca generated accepted values and criteria such as clarity, efficiency, and more - see \tab~\ref{tab:criteria-math}.

\item \qa: The goal of \qa is to quantify each of the suggested criterion, to access the task utility of the system $U_t$, for the end user. We define the Utility for task $t$ as: \utility. where $s$ represents the task sample and $Q(s|c_i.)$ is the quantifier output for sample $s$ based on the criterion $c_i$. For example, for math problem solving, given the generated criteria shown in \tab~\ref{tab:criteria-math}, the solution's Accuracy could be quantified as ``Incorrect'', ``partially correct'' or ``correct''. Eligible quantified values for quantification process are shown in ``Accepted values'' column in \tab~\ref{tab:criteria-math}

\item \va: There might be cases where not all the criteria suggested by \ca help assess utility. Some criteria might be redundant, while others may not aid in distinguishing performance. \va validates the quality of the criteria in terms of robustness and their distinguishability of noisy samples. Essentially, it checks (1) if the criteria can be quantified robustly over repeated samples, and (2) if \qa can identify the adversarial attacked targeted samples from the original ones. If the sanity checks do not pass, \va will update the list of criteria, to end up with a set of robust, stable, informative and distinguishable criteria for assessment.
\end{itemize}

\noindent
Finally, we note that \emph{AgentEval} allows for incorporating a human in the loop in the role of a domain expert. For instance, \ca could be replaced by a human expert who either comes up with the relevant criteria or helps \va verify the useful criteria and filter out the unessential ones.
\section{Datasets and Solutions}
\label{sec:datasets}
This section provides an overview of the datasets utilized in our study i.e., Math problem solving and ALFWorld household task. 
The math dataset is chosen for its widespread usage and complex problem-solving scenarios that are fundamental in evaluating the effectiveness.
ALFWorld dataset offers a scenario involving multi-turn interactions within a moderately approximated multi-modal environment. 
Each dataset plays a critical role in evaluating different aspects of \emph{AgentEval}'s capabilities, from handling complex theoretical problems to navigating real-world scenarios. 
In both tasks, although success is clearly defined, multiple solutions exist for accomplishing the objectives. 
An example of Math problem solving and ALFWorld task is shown in Appendix~\ref{sec:task-example}.
Due to space, we report all experiments about Math problem solving in the main paper and we keep all the experiments related to ALFWorld dataset in the Appendix~\ref{apd:alf}.

\subsection{MATH Problem Solving}
\label{sec:math-problems}
\textbf{Dataset:} The MATH dataset is a substantial collection of 12,500 challenging mathematics problems from high school competitions~\cite{hendrycks2021measuring}. Each problem comes with a step-by-step solution and is tagged by difficulty levels. 
Similar to the math problem experimental setup in \citet{wu2023autogen}, we carry out evaluations on 120 problems from level-5 by three different solutions. Due to limited space, for more details about this dataset, we refer readers to Appendix~\ref{apd:math}

\noindent
\textbf{Solutions:} In establishing solutions for this task to assess, we draw inspiration from the experiments showcased in \cite{wu2023autogen}. We evaluate the proposed methodology by AutoGen~\cite{wu2023autogen}, as well as Langchain ReAct~\cite{yao2022react} 
and a Vanilla solver that employs GPT-4 to tackle the task. These solutions have previously demonstrated promising and competitive performance~\cite{wu2023autogen}.
In Sec.~\ref{sec:agenteval-math}, we explore how the measured performance with \emph{AgentEval} correlates with the ground truths.

\subsection{ALFWorld Household Task}
\label{sec:alfworld}

\textbf{Dataset:} ALFWorld presents a set of language-based interactive decision-making tasks within simulated household environments \cite{shridhar2020alfworld}. 
ALFWorld is the first interactive parallel environment that aligns text descriptions and commands with physically embodied robotic simulation. 
Finally, the dataset's inclusion of household chores to more intricate problem-solving scenarios, provides a comprehensive testbed for evaluating the adaptability of multi-agent systems. For more information about the dataset and examples of the test cases, we refer the readers to Appendix~\ref{apd:alf-data}.

\noindent
\textbf{Solutions:}
As for the solutions to assess for ALFWorld Household tasks, similar to \cite{wu2023autogen}, we consider ReAct~\cite{yao2022react} as well as AutoGen with two agents and AutoGen with three agents \cite{wu2023autogen}. In Appendix~\ref{apd:alf-sol}, we discuss in more details the solutions under assessment. 
We assess and compare the performance of these three solutions using \emph{AgentEval}.

\section{Experiments}
\label{sec:workflow}

\subsection{Implementation Details}
For all experiments, we use GPT-4 version 0613, accessed through Azure OpenAI services, as the LLM model and the temperature of 0. 
\emph{AgentEval} utilizes AutoGen~\cite{wu2023autogen}
for implementation, since it provides a versatile environment where agents can be finely tuned and customized based on specific application needs. This is crucial for maintaining the flexibility to handle a wide range of applications. We tried to avoid much prompt engineering and tried to keep each agent's instructions as if we are instructing human annotators. Moreover, another advantages of using AutoGen for implementation of \emph{AgentEval} is that it has the flexibility to involve human in the loop. Each agent could be replaced by a human annotator. We further provide all the prompts used in our experiments in our Git repository.

\subsection{AgentEval for Math Problems}
\label{sec:agenteval-math}
\begin{figure}[]
\vspace{-1em}
  \includegraphics[clip, trim=7cm 1.2cm 7cm 7.9cm,scale=0.73]{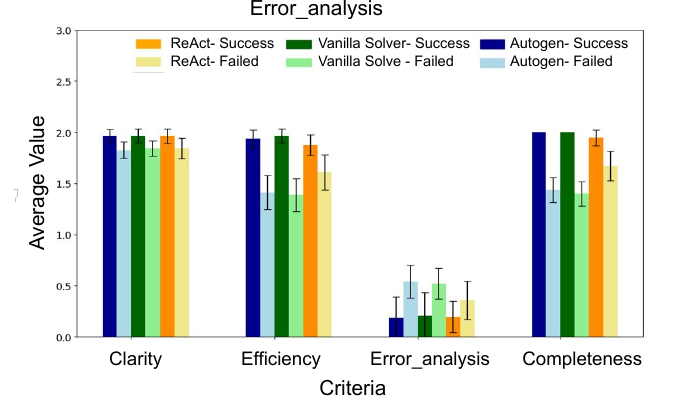}
\caption{\emph{AgentEval} assessment of three  solutions on math problems categorized by success and failed cases.}
\vspace{-1em}
\label{img:math-exp1}
\end{figure}

When executing the \ca for Math problem solving, we first obtain a set of criteria as presented in Tab.~\ref{tab:criteria-math}. Then, the \qa is tasked with quantifying each criterion, based on the accepted values. We present the outcome of \qa measuring performance of three solutions on this task in \fig~\ref{img:math-exp1}. Notably, we see that \emph{Agenteval} does not quantify the three solutions as if they perform equally well across the different criteria. For instance, while all three solutions leverage GPT-4 as the underlying language model, Autogen outperforms ReAct and Vanilla GPT-4 in terms of accuracy. This observation, while confirmed by previous studies~\cite{wu2023autogen}, extends to solution completeness and efficiency as well.  

\begin{figure*}[!t]
\centering\vspace{-2.5em}
\includegraphics[width=1\textwidth]{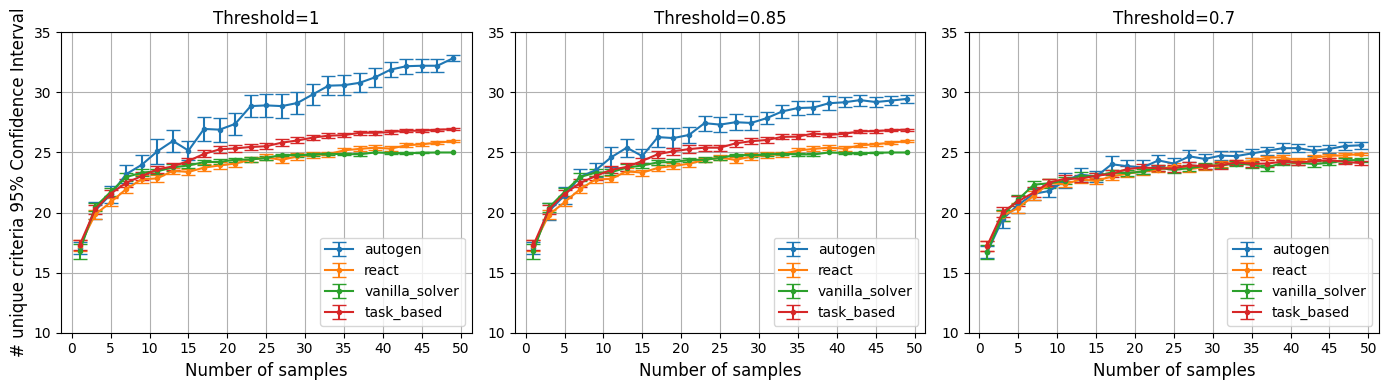}
\vspace{-0.5em}
\caption{Task-based vs solution-based criteria for Math problems. Error bar show the 95\% confidence interval.}
\vspace{-1em}
\label{fig:task-vs-solution}
\end{figure*}

As depicted in \fig~\ref{img:math-exp1}, the error analysis range of quantified values differs from other metrics. We scrutinize the results by categorizing them into successful and failed cases.
AutoGen, Vanilla Solver and ReAct solutions are each presented in orange, blue and green respectively, where the darker bars represent the performance on successful cases and lighter bars represent the failed cases. The difference between the dark and light bar of each color, verify AgentEval's performance, as we expect that each positive criteria should be quantified higher for successful cases compared to their failed cases. We observe that in most cases, the successful and failed cases are distinguished, even with 95\% interval confidence on all the success and failed cases. 

When examining the differences between successful and failed cases among the three solutions, we note that not all successful cases are assessed identically, nor are all failed cases quantified with the same performance. This can be interpreted to mean that even though two solutions might both be successful, one might perform better or worse in certain criteria, such as clarity or efficiency. This observation provides us with valuable additional insights, especially for the developers of the proposed solutions, and goes beyond reporting the effectiveness of a application by one scalar value e.g., success rate.


\begin{table}[t]
    \centering
    \vspace{-1em}
        \caption{Verification Criteria for MathProblems}

    \scalebox{0.56}{
    \begin{tabular}{>{\raggedright\arraybackslash}p{1.9cm}>{\raggedright\arraybackslash}p{6.3cm}>{\raggedright\arraybackslash}p{4.5cm}}
        \hline \hline
        \textbf{Criteria} & \textbf{Description} & \begin{center}
        \vspace{-1em}
            \textbf{Accepted Values}
                    \vspace{-1em}
        \end{center}  \\

        \hline
       Clarity & The ease of understanding the steps, explanations, and language used in the solution. & 
                                       \vspace{-0.5em}
 \begin{enumerate}[label=--,itemsep=0em,parsep=-0.2em,topsep=-4em]
            \item Not Clear   (0)
            \item Moderately Clear (1)
            \item Very Clear (2)
        \end{enumerate} 
       \\[-1em]
        \hline
        Efficiency & The use of optimal methods or approaches to solve the math problem. & 
                                       \vspace{-0.5em}
 \begin{enumerate}[label=--,itemsep=0em,parsep=-0.2em,topsep=-4em]
            \item  Inefficient  (0)
            \item Moderately Efficient (1)
            \item Efficient (2)
 
        \end{enumerate}
    \\[-1em]
        \hline
        Error Analysis & The identification and description of possible errors or misconceptions in the math problem-solving process. &              
                                       \vspace{-0.5em}
 \begin{enumerate}[label=--,itemsep=0em,parsep=-0.2em,topsep=-4em]
            \item  Not Addressed  (0)
            \item  Partially Addressed (1)
            \item Well Addressed (2)
 
        \end{enumerate}
    \\[-1em]
        \hline
        Completeness & Quality of code in terms of efficiency and elegance & 
                                       \vspace{-0.5em}
 \begin{enumerate}[label=--,itemsep=0em,parsep=-0.1em,topsep=-4em]
            \item Incomplete  (0)
            \item Mostly Complete (1)
            \item Complete (2)
        \end{enumerate}
        \\[-1em]
        \hline \hline

    \end{tabular}
    }
    \label{tab:criteria-math}
\end{table}

\section{Robustness Analysis and Verification}
\label{sec:results}
In this section, we first analyze the robustness of \emph{AgentEval}, then further investigate how \va can increase the stability of our assessment.

\subsection{Diversity of Criteria}
\label{sec:c-robust}
Here, our main goal is to study the diversity of the suggested criteria. 
We investigate the extent inputs to \emph{AgentEval} (\fig~\ref{fig:agent-eval} such as `Task Description' and `Successful/Failed Executions') contribute to \ca for creating a more diverse set of criteria. To do so, we use two distinct methods, with \ca generating (1) ``task-based'' criteria solely from the task description, and (2) ``solution-based'' criteria, derived from both the task and execution examples.  For example, a solution to a mathematical problem, might satisfy criteria such as `Accuracy' and `Clarity', independent of the solution. However, when additional tools such as coding are used to solve the problems, additional criteria like `Code Efficiency' may be introduced to the set of criteria. This makes sense, since the application leveraged coding to solve math problems. 
 
Fig.~\ref{fig:task-vs-solution} displays the number of unique criteria extracted for mathematical problem solving in task-based mode, and three different solution-based approaches.
To keep the balance between computational costs and analyzing the robustness, we conducted 50 runs of the \ca with different seeds. Subsequently, for $N=50$ iterations, we randomly select $M \le 50$ samples, as shown on the x-axis of Fig.~\ref{fig:task-vs-solution}, and present the average number of unique extracted criteria, along with its 95\% confidence interval after repeating this process 50 times. We note that because the  total pool of criteria includes 50 iterations in total, the confidence intervals become smaller when $M$ get closer to the
maximum number of samples i.e., 50

To gain deeper insights into diversity of criteria, we took a closer look at them to study if they are truly unique or to what extent they have similarities. This is important to determine if \ca, when continually generating criteria, will always produce new criteria, or if it will eventually converge to a set.
We noted that some criteria are similar but worded differently. For example, `Problem Complexity' vs. `Problem Difficulty' or `Time Taken' vs. `Time to Completion'. Tab.~\ref{tab:sim_criteria} in the Appendix lists such instances.
To consolidate the similar criteria and reduce noise in the number of unique criteria and redundancy, inspired from previous work \cite{liu2022semantic,vahtola2022not,reimers2019sentence}, we employ a pre-trained language model fine-tuned for paraphrasing\footnote{\url{https://bit.ly/3UgsYOp}}, to measure the semantic similarity of criteria descriptions. Using a threshold $\tau$, we classify pairs with cosine similarity greater than $\tau$ as semi-identical ones and select one of them as the representative of the pair. 
\fig~\ref{fig:task-vs-solution} illustrates the impact of different $\tau$ values (0.7, 0.85, 1) on the diversity of criteria. A threshold of 1 means no filtering occurs.
This analysis shows that the solution-based approach has potential to produce more diverse criteria than the task-based approach, although this varies by the creativity of the model. For example, while the AutoGen solution demonstrates the highest diversity, task-based methods yield more unique criteria than ReAct and Vanilla Solver. Another interesting observation is that repeating the \ca will eventually lead to a convergence in the number of criteria. This suggests that the 
\ca's ability to create new criteria will diminish, converging to an almost finite list of criteria, which will reduce the cost as well.

\subsection{Verification}
\label{sec:verify}

As outlined in Sec.~\ref{sec:agenteval} and illustrated in Fig.~\ref{fig:agent-eval}, the \va's primary role is to ensure the selected criteria are effective toward evaluating the utility for the end-user, while maintaining robustness and high discriminative power. To achieve this, the \va undertakes two main actions:

\noindent    
\textit{(1) Criteria Stability:} 
The criteria should be essential and robust, meaning they should not be redundant and we should be able to quantify them stably if we repeatedly quantify it for an individual solution, showing no divergence.
As such, \va  enhances the criteria by iterating over the generation and quantification phases. It then consolidates these criteria by identifying and eliminating redundancies, followed by evaluating the dispersion of the distribution of the quantified criteria. 
This step modifies the criteria, ensuring that only the most robust criteria are retained.

\noindent \textit{(2) Discriminative Power:} A reliable evaluation should detect and withstand noise. To test that, we propose to use adversarial examples and then assess the system's ability to differentiate between these compromised examples and standard cases. Should the system fail to distinguish effectively, it indicates that the criteria are insufficient for reliable assessment under varied conditions.

We note that both steps involve a tunable threshold that can be adapted based on application needs, ensuring flexible criteria validation.  The proposed methodology for \va is summarized in Algorithm~\ref{alg:verifier} in the Appendix.

\subsubsection{Criteria Stability}
\label{sec:q-robust}

\begin{figure}[!tb]
\centering
\vspace{-2em}
\includegraphics[width=0.47\textwidth]{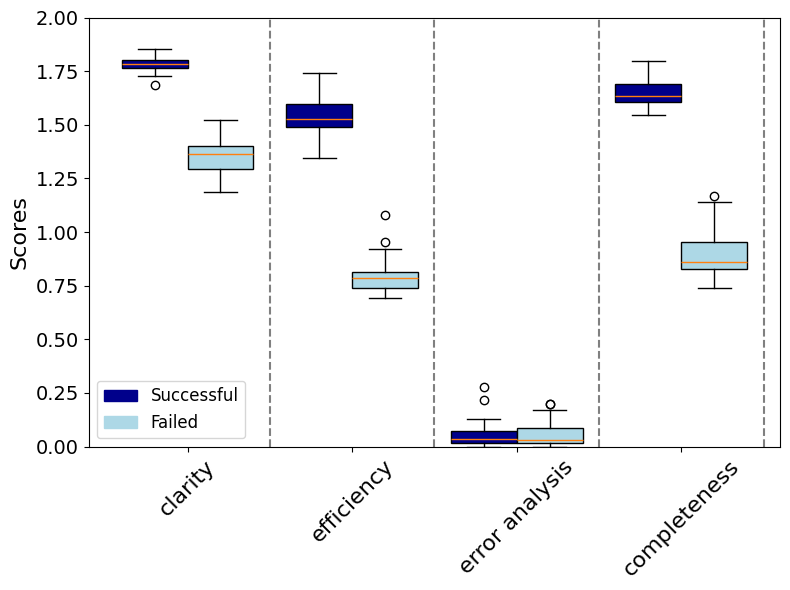}
\caption{Distribution of \qa output on AutoGen results on successful (dark blue) and failed (light blue) cases on different criteria.}
\vspace{-1em}
\label{fig:q-box-plot}
\end{figure}

Our goal here is to explore the stability of criteria and robustness of the quantifier for having a more essential, robust and stable set of criteria. 
We specifically evaluate the \qa's robustness using criteria for mathematical problems (Table \ref{tab:criteria-math}), conducting 50 repeats of runs with different seeds on 120 problems (Section \ref{sec:math-problems}). Ideal expected outcomes include consistent performance across all criteria on all the repeats.
\fig~\ref{fig:q-box-plot} illustrates the distribution of quantifier values for both failed (dark blue) and successful cases (light blue) across all criteria through box plots. The more robust a criterion, the narrower the range of quantified performance (narrower box plots). Also, the less overlap between the successful and failed boxes, the higher the distinguishability of the criteria.
We observe that  all four criteria, except `error analysis' allow for easy differentiation between successful and failed cases. Additionally, some criteria prove to be more robust compared to others. We believe that such an analysis of the quantifier agent's performance will yield valuable insights for enhancing reliability, trustworthiness, and explainability in performance evaluation. A detailed examination of the stability of each criterion, especially how they differentiate between successful and failed cases, is provided in Appendix~\ref{apd:robustness}.

Further, to refine and expand the criteria set without redundancy, we operate the \ca multiple times i.e., we execute \ca 50 times with varied seeds. The criteria are then summarized into one list of useful criteria using the LLM. Additionally, as explained in Section \ref{sec:c-robust}, we  remove similar and redundant criteria using pre-trained language models, thus obtaining a comprehensive list of criteria. 
The refined criteria after 50 repeats are detailed in Tab.~\ref{tab:full-criteria-math} in the Appendix.

\begin{figure}[t]
\centering
\vspace{-2em}
  \includegraphics[clip, trim=0cm 0cm 0cm 0cm,scale=0.34]{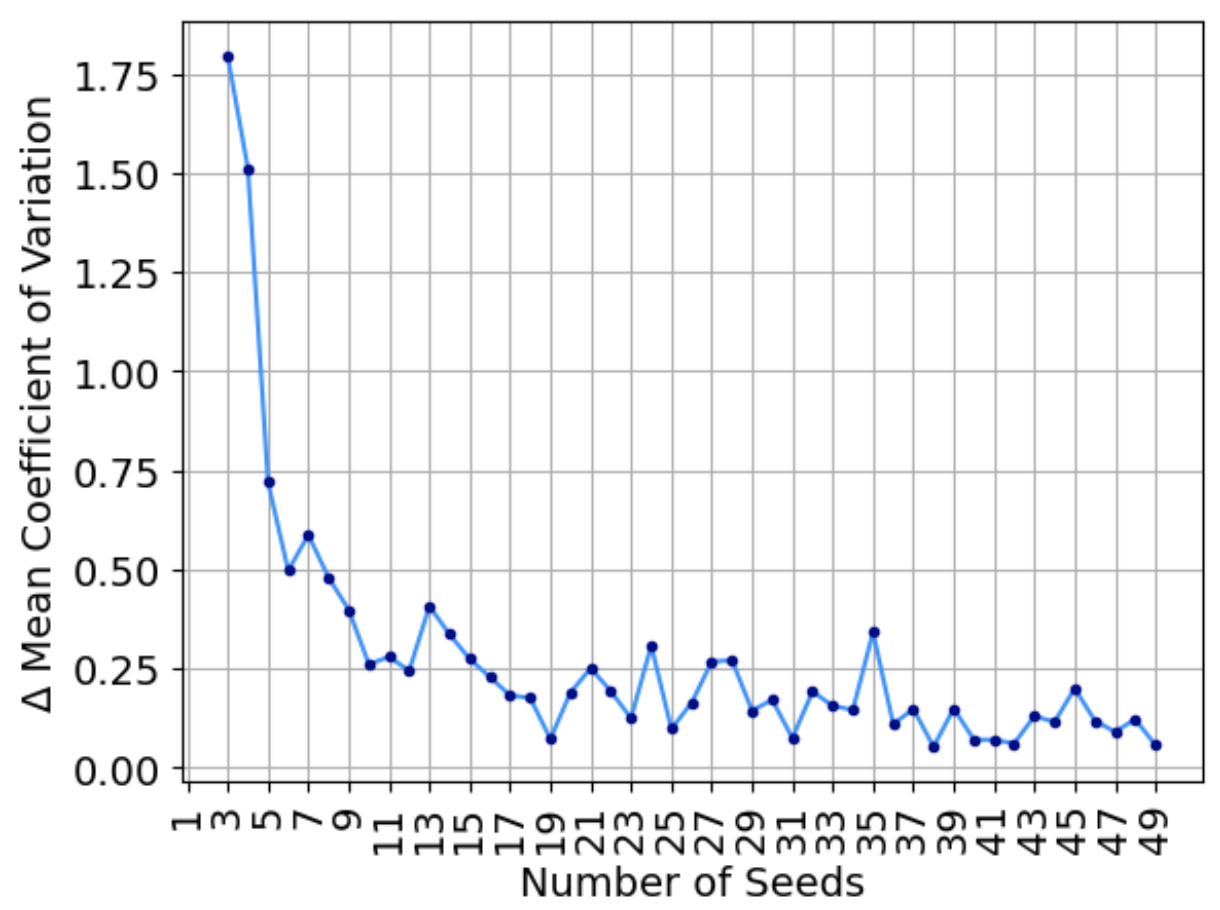}
\caption{$\Delta$ sum of mean coefficient of variation across all criteria with increasing number of seeds.}
\vspace{-1em}
\label{fig:num_seeds}
\end{figure}

Now, we aim to determine the stability of these criteria through repeated quantifications. Our goal is to identify criteria that maintain consistent results without significant divergence, even when quantified multiple times.
Using this consolidated list, we measure the dispersion of quantified results using the coefficient of variation, a standardized metric that facilitates comparison across various test cases when \qa quantifies them. 
Given the consolidated list of criteria, we use the  \qa to quantify various test cases and report the coefficient of variation as a measure of the dispersion of the \qa's outputs with respect to each criterion across different seeds  and report the mean coefficient of variation across all samples.   
we run \qa with 50 seeds and plot the change ($\Delta$) in the sum of mean coefficient of variation across all criteria against the number of seeds, in Figure~\ref{fig:num_seeds}. For each criterion, we compute the absolute difference with the mean coefficient of variation calculated when using $n-1$ seeds, summing up the absolute differences across all criteria. According to the plot, after approximately 18 seeds, the magnitude of mean coefficient of variation stabilizes and becomes rather trivial.
In almost all cases, the mean coefficient of variation is around or below 0.5, which is relatively small, suggesting that \qa is quite robust.

\subsubsection{Discriminative Power}
\label{sec:q-verify}
\begin{figure}[t]
\centering
\includegraphics[width=0.5\textwidth]{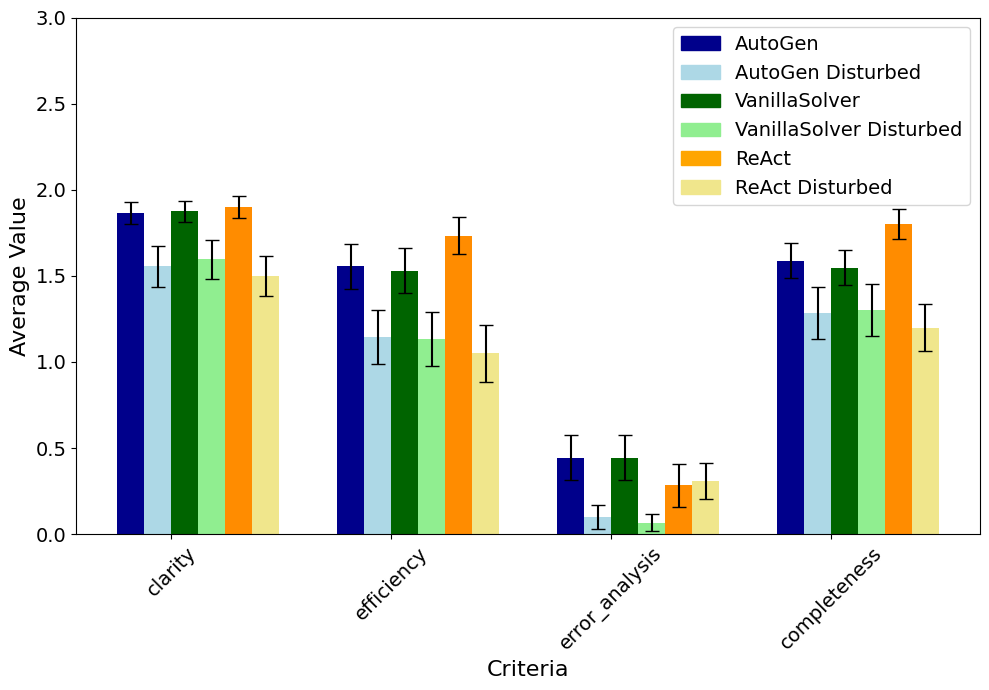}
\caption{
Assessment of original and disturbed solutions on Math dataset (discriminative power study).}
\vspace{-1em}
\label{fig:q-verify-disturb}
\end{figure}

It is crucial to ensure the quality of quantification of each criterion. Ideally, this validation would involve comparisons with known pairwise samples, where sample $S^{+}$  is definitively superior to $S^{-}$  for a given criterion. If the evaluator also confirms superiority of $S^{+}$ w.r.t $S^{-}$, it has robust quantification. However, due to rapid expansion of LLM-powered applications, obtaining annotated data for many tasks is often unfeasible. Therefore, we propose using synthetically altered versions of samples for verification.
Let us assume we have an alternative disturbed version of sample $S$, which is called $S'$. 
Assuming sample $S$ is more likely to outperform its disturbed version $S'$, our assessment should confirm this assumption by assigning better quantified performance $S$ in comparison to $S'$. In experiments with mathematical problems, we introduced random noise by removing portions of the solution sentences from AutoGen, VanillaSolver, and ReAct's results respectively, expecting that criteria like `Completeness' or `Clarity' would show be higherin $S$ than in $S'$.
 We disturbed solutions by removing 25\% of the sentences and assessed the \qa's performance. As shown in Fig.~\ref{fig:q-verify-disturb}, criteria measuring  aspects like `Clarity' and `Completeness' were lower in disturbed solutions (lighter bars), confirming \qa's high discriminative power and effectiveness.

We have already filtered out the criteria that were unstable, i.e., those that had a high mean standard deviation and dispersion when being quantified in the previous section. We report the results of the \qa quantifying differences between original and disturbed samples on the comprehensive set of criteria shown in Appendix, as shown in Fig.~\ref{fig:adversarial_results} for the math problem-solving. In most cases, the \qa quantifies the disturbed output to be worse than the original task output. We believe analyzing the \qa's performance will enhance the reliability, trustworthiness, and explainability in evaluations..

\subsubsection{\va}
After modifying the list of criteria (Sec.~\ref{sec:q-robust}), we have developed a stable and robust list of criteria that the \qa can reliably quantify. Further, we also proposed a method for assessing whether the criteria can distinguish between noise-adversarially attacked samples and the original ones. 
These two tests will serve as input for the \va (described in Algorithm \ref{alg:verifier}), which can also have its threshold tuned for different applications. For instance, one might prioritize the stability of the criteria, while another may value the discriminative power of the \emph{AgentEval} for specific applications. As such, the \va will modify and update the criteria based on to what extend they pass the two tests, i.e., if the mean coefficient of variation is below a specific threshold and the percentage of adversarial testing it has passed. The \va will then update the criteria if necessary.  We believe that having a \va would help continuously updating the criteria as needed because, by improving the systems, we may require new criteria that were not previously necessary for utility assessment.

\section{Conclusions and Future Work}
\label{sec:conclusions}


We introduced the \emph{AgentEval} framework, designed to swiftly gauge the utility of arbitrary LLM-powered agentic applications. Our framework leverages recent findings suggesting LLMs as a scalable and cost-effective alternative to human evaluations for open-ended tasks. \emph{AgentEval} consists of three agents: \ca suggests criteria based on task descriptions and executions of the applications, \qa quantifies how well the application flow aligns with these criteria, and \va modifies the list of criteria if needed. This framework is customizable, adaptable, and can operate in various modes, employing combinations of LLMs, human inputs, and tools.
We believe that suggested \emph{AgentEval}'s utility extends beyond immediate performance. It can uncover new system capabilities over time and adapt to changes in user needs tracked by developers. \emph{AgentEval} can also enable developers to assess the alignment between application behavior and suggested user requirements, providing them with insights into areas for improvement.
In summary, our contributions include introducing the \emph{AgentEval} framework, and conducting a robust analysis of its performance across various datasets and baselines. \emph{AgentEval} represents a significant step towards assessing LLM-powered applications.

\section{Limitations and Ethics}
\label{sec:limitation}

\subsection{Limitations}
Here, we discuss some limitations of the \emph{AgentEval} framework. Firstly, the performance of the AgentEval is highly dependent on the quality of the output logs of the applications. Flaws or limitations in these outputs can significantly impact the framework's ability to accurately assess utility.

Secondly, our experiments were conducted exclusively with closed-source LLMs, specifically with GPT-4. This may limit the generalizability of our findings. Plans to include a broader array of LLMs, including open-source models, are considered for future studies to validate and possibly enhance the robustness of our conclusions.
Additionally, the tests conducted were limited to specific scenarios within math problem solving and household tasks. Expanding the diversity of test scenarios could help in understanding the broader applicability of the framework.

Thirdly, while \emph{AgentEval} employs a novel methodology leveraging LLMs to estimate utility, the absence of human evaluation in our validation process could be viewed as a drawback. Human evaluations provide unique insights, especially in subjective aspects of utility that automated systems might overlook. However, such evaluations are often cost-prohibitive and logistically challenging, restricting our ability to implement them within this study. Especially do developers of agentic LLM-powered applications who needs insights fast as they go with the deployments.

Lastly, as LLM technologies evolve, the criteria and metrics used for evaluation may need to be updated or revised. What works for assessing current LLMs may not hold as these models become more advanced. Continuous updates to the evaluation framework will be necessary to keep pace with technological advancements.

\subsection{Ethics}
To the best of our knowledge, we did not violate any code of ethics with the experiments done in this paper. We reported technical details and results, with details in the main paper, Appendix, and code release. Our experimental results are an outcome of a Machine Learning model.

Our \emph{AgentEval} system has a variety of uses in real world settings, such as improving applications for end users or helping developers. However, we caution that it must be used carefully, as the outputs are from a ML model and can have real world consequences, if used incorrectly.

These and many other related issues are important aspects to consider when deploying a system like \emph{AgentEval} in the real world.
\section*{Acknowledgement}
We thank Besmira Nushi, Victor Dibia, and Adam Fourney for inspiring discussions for earlier versions of the AgentEval framework.

\bibliography{acl}

\begin{thebibliography}{62}
\expandafter\ifx\csname natexlab\endcsname\relax\def\natexlab#1{#1}\fi

\bibitem[{Ahmadvand et~al.(2022)Ahmadvand, Arabzadeh, Kiseleva, Sanz, Deng, Jauhar, Gamon, Agichtein, Friend et~al.}]{ahmadvand2022supporting}
Ali Ahmadvand, Negar Arabzadeh, Julia Kiseleva, Patricio~Figueroa Sanz, Xin Deng, Sujay Jauhar, Michael Gamon, Eugene Agichtein, Ned Friend, et~al. 2022.
\newblock Supporting complex information-seeking tasks with implicit constraints.
\newblock \emph{arXiv preprint arXiv:2205.00584}.

\bibitem[{Andrew(2024)}]{andrew2024potential}
Albert Andrew. 2024.
\newblock Potential applications and implications of large language models in primary care.
\newblock \emph{Family Medicine and Community Health}, 12(Suppl 1).

\bibitem[{Azzopardi et~al.(2018)Azzopardi, Thomas, and Craswell}]{azzopardi2018measuring}
Leif Azzopardi, Paul Thomas, and Nick Craswell. 2018.
\newblock Measuring the utility of search engine result pages: an information foraging based measure.
\newblock In \emph{The 41st International ACM SIGIR conference on research \& development in information retrieval}, pages 605--614.

\bibitem[{Bang et~al.(2023)Bang, Cahyawijaya, Lee, Dai, Su, Wilie, Lovenia, Ji, Yu, Chung, Do, Xu, and Fung}]{bang2023multitask}
Yejin Bang, Samuel Cahyawijaya, Nayeon Lee, Wenliang Dai, Dan Su, Bryan Wilie, Holy Lovenia, Ziwei Ji, Tiezheng Yu, Willy Chung, Quyet~V. Do, Yan Xu, and Pascale Fung. 2023.
\newblock \href {http://arxiv.org/abs/2302.04023} {A multitask, multilingual, multimodal evaluation of chatgpt on reasoning, hallucination, and interactivity}.

\bibitem[{Bano et~al.(2023)Bano, Zowghi, and Whittle}]{bano2023exploring}
Muneera Bano, Didar Zowghi, and Jon Whittle. 2023.
\newblock Exploring qualitative research using llms.
\newblock \emph{arXiv preprint arXiv:2306.13298}.

\bibitem[{Bian et~al.(2023)Bian, Han, Sun, Lin, Lu, and He}]{bian2023chatgpt}
Ning Bian, Xianpei Han, Le~Sun, Hongyu Lin, Yaojie Lu, and Ben He. 2023.
\newblock Chatgpt is a knowledgeable but inexperienced solver: An investigation of commonsense problem in large language models.
\newblock \emph{arXiv preprint arXiv:2303.16421}.

\bibitem[{Borisov et~al.(2018)Borisov, Kiseleva, Markov, and de~Rijke}]{borisov2018calibration}
Alexey Borisov, Julia Kiseleva, Ilya Markov, and Maarten de~Rijke. 2018.
\newblock Calibration: A simple way to improve click models.
\newblock In \emph{Proceedings of the 27th ACM International Conference on Information and Knowledge Management}, pages 1503--1506.

\bibitem[{Chan et~al.(2023)Chan, Chen, Su, Yu, Xue, Zhang, Fu, and Liu}]{chan2023chateval}
Chi-Min Chan, Weize Chen, Yusheng Su, Jianxuan Yu, Wei Xue, Shanghang Zhang, Jie Fu, and Zhiyuan Liu. 2023.
\newblock Chateval: Towards better llm-based evaluators through multi-agent debate.
\newblock In \emph{The Twelfth International Conference on Learning Representations}.

\bibitem[{Chang et~al.(2023)Chang, Wang, Wang, Wu, Yang, Zhu, Chen, Yi, Wang, Wang et~al.}]{chang2023survey}
Yupeng Chang, Xu~Wang, Jindong Wang, Yuan Wu, Linyi Yang, Kaijie Zhu, Hao Chen, Xiaoyuan Yi, Cunxiang Wang, Yidong Wang, et~al. 2023.
\newblock A survey on evaluation of large language models.
\newblock \emph{ACM Transactions on Intelligent Systems and Technology}.

\bibitem[{Chern et~al.(2024)Chern, Chern, Neubig, and Liu}]{chern2024can}
Steffi Chern, Ethan Chern, Graham Neubig, and Pengfei Liu. 2024.
\newblock Can large language models be trusted for evaluation? scalable meta-evaluation of llms as evaluators via agent debate.
\newblock \emph{arXiv preprint arXiv:2401.16788}.

\bibitem[{Chiang and Lee(2023)}]{chiang2023can}
Cheng-Han Chiang and Hung-yi Lee. 2023.
\newblock Can large language models be an alternative to human evaluations?
\newblock \emph{arXiv preprint arXiv:2305.01937}.

\bibitem[{C{\^o}t{\'e} et~al.(2019)C{\^o}t{\'e}, K{\'a}d{\'a}r, Yuan, Kybartas, Barnes, Fine, Moore, Hausknecht, El~Asri, Adada et~al.}]{cote2019textworld}
Marc-Alexandre C{\^o}t{\'e}, Akos K{\'a}d{\'a}r, Xingdi Yuan, Ben Kybartas, Tavian Barnes, Emery Fine, James Moore, Matthew Hausknecht, Layla El~Asri, Mahmoud Adada, et~al. 2019.
\newblock Textworld: A learning environment for text-based games.
\newblock In \emph{Computer Games: 7th Workshop, CGW 2018, Held in Conjunction with the 27th International Conference on Artificial Intelligence, IJCAI 2018, Stockholm, Sweden, July 13, 2018, Revised Selected Papers 7}, pages 41--75. Springer.

\bibitem[{Dibia et~al.(2023)Dibia, Fourney, Bansal, Poursabzi-Sangdeh, Liu, and Amershi}]{dibia2023aligning}
Victor Dibia, Adam Fourney, Gagan Bansal, Forough Poursabzi-Sangdeh, Han Liu, and Saleema Amershi. 2023.
\newblock \href {http://arxiv.org/abs/2210.16494} {Aligning offline metrics and human judgments of value for code generation models}.

\bibitem[{Fu et~al.(2023)Fu, Ng, Jiang, and Liu}]{fu2023gptscore}
Jinlan Fu, See-Kiong Ng, Zhengbao Jiang, and Pengfei Liu. 2023.
\newblock Gptscore: Evaluate as you desire.
\newblock \emph{arXiv preprint arXiv:2302.04166}.

\bibitem[{Guo et~al.(2023)Guo, Jin, Liu, Huang, Shi, Yu, Liu, Li, Xiong, Xiong et~al.}]{guo2023evaluating}
Zishan Guo, Renren Jin, Chuang Liu, Yufei Huang, Dan Shi, Linhao Yu, Yan Liu, Jiaxuan Li, Bojian Xiong, Deyi Xiong, et~al. 2023.
\newblock Evaluating large language models: A comprehensive survey.
\newblock \emph{arXiv preprint arXiv:2310.19736}.

\bibitem[{Hendrycks et~al.(2021{\natexlab{a}})Hendrycks, Burns, Basart, Critch, Li, Song, and Steinhardt}]{hendrycks2020aligning}
Dan Hendrycks, Collin Burns, Steven Basart, Andrew Critch, Jerry Li, Dawn Song, and Jacob Steinhardt. 2021{\natexlab{a}}.
\newblock Aligning ai with shared human values.
\newblock \emph{Proceedings of the International Conference on Learning Representations (ICLR)}.

\bibitem[{Hendrycks et~al.(2021{\natexlab{b}})Hendrycks, Burns, Kadavath, Arora, Basart, Tang, Song, and Steinhardt}]{hendrycks2021measuring}
Dan Hendrycks, Collin Burns, Saurav Kadavath, Akul Arora, Steven Basart, Eric Tang, Dawn Song, and Jacob Steinhardt. 2021{\natexlab{b}}.
\newblock Measuring mathematical problem solving with the math dataset.
\newblock \emph{arXiv preprint arXiv:2103.03874}.

\bibitem[{Hong et~al.(2023)Hong, Zheng, Chen, Cheng, Wang, Zhang, Wang, Yau, Lin, Zhou et~al.}]{hong2023metagpt}
Sirui Hong, Xiawu Zheng, Jonathan Chen, Yuheng Cheng, Jinlin Wang, Ceyao Zhang, Zili Wang, Steven Ka~Shing Yau, Zijuan Lin, Liyang Zhou, et~al. 2023.
\newblock Metagpt: Meta programming for multi-agent collaborative framework.
\newblock \emph{arXiv preprint arXiv:2308.00352}.

\bibitem[{Huang et~al.(2023)Huang, Chen, Mishra, Zheng, Yu, Song, and Zhou}]{huang2023large}
Jie Huang, Xinyun Chen, Swaroop Mishra, Huaixiu~Steven Zheng, Adams~Wei Yu, Xinying Song, and Denny Zhou. 2023.
\newblock Large language models cannot self-correct reasoning yet.
\newblock In \emph{The Twelfth International Conference on Learning Representations}.

\bibitem[{Jain et~al.(2023)Jain, Keshava, Sathyendra, Fernandes, Liu, Neubig, and Zhou}]{jain2023multi}
Sameer Jain, Vaishakh Keshava, Swarnashree~Mysore Sathyendra, Patrick Fernandes, Pengfei Liu, Graham Neubig, and Chunting Zhou. 2023.
\newblock Multi-dimensional evaluation of text summarization with in-context learning.
\newblock \emph{arXiv preprint arXiv:2306.01200}.

\bibitem[{Jin et~al.(2019)Jin, Dhingra, Liu, Cohen, and Lu}]{jin2019pubmedqa}
Qiao Jin, Bhuwan Dhingra, Zhengping Liu, William~W Cohen, and Xinghua Lu. 2019.
\newblock Pubmedqa: A dataset for biomedical research question answering.
\newblock \emph{arXiv preprint arXiv:1909.06146}.

\bibitem[{Kiseleva et~al.(2014)Kiseleva, Crestan, Brigo, and Dittel}]{kiseleva2014modelling}
Julia Kiseleva, Eric Crestan, Riccardo Brigo, and Roland Dittel. 2014.
\newblock Modelling and detecting changes in user satisfaction.
\newblock In \emph{Proceedings of the 23rd ACM International Conference on Conference on Information and Knowledge Management}, pages 1449--1458.

\bibitem[{Kiseleva and de~Rijke(2017)}]{kiseleva2017evaluating}
Julia Kiseleva and Maarten de~Rijke. 2017.
\newblock Evaluating personal assistants on mobile devices.
\newblock \emph{arXiv preprint arXiv:1706.04524}.

\bibitem[{Kiseleva et~al.(2022{\natexlab{a}})Kiseleva, Li, Aliannejadi, Mohanty, ter Hoeve, Burtsev, Skrynnik, Zholus, Panov, Srinet, Szlam, Sun, Hofmann, C{\^o}t{\'e}, Awadallah, Abdrazakov, Churin, Manggala, Naszadi, van~der Meer, and Kim}]{pmlr-v176-kiseleva22a}
Julia Kiseleva, Ziming Li, Mohammad Aliannejadi, Shrestha Mohanty, Maartje ter Hoeve, Mikhail Burtsev, Alexey Skrynnik, Artem Zholus, Aleksandr Panov, Kavya Srinet, Arthur Szlam, Yuxuan Sun, Katja Hofmann, Marc-Alexandre C{\^o}t{\'e}, Ahmed Awadallah, Linar Abdrazakov, Igor Churin, Putra Manggala, Kata Naszadi, Michiel van~der Meer, and Taewoon Kim. 2022{\natexlab{a}}.
\newblock \href {https://proceedings.mlr.press/v176/kiseleva22a.html} {Interactive grounded language understanding in a collaborative environment: Iglu 2021}.
\newblock In \emph{Proceedings of the NeurIPS 2021 Competitions and Demonstrations Track}, volume 176 of \emph{Proceedings of Machine Learning Research}, pages 146--161. PMLR.

\bibitem[{Kiseleva et~al.(2022{\natexlab{b}})Kiseleva, Skrynnik, Zholus, Mohanty, Arabzadeh, C\^{o}t\'e, Aliannejadi, Teruel, Li, Burtsev, ter Hoeve, Volovikova, Panov, Sun, Srinet, Szlam, Awadallah, Rho, Kwon, Wontae~Nam, Bivort~Haiek, Zhang, Abdrazakov, Qingyam, Zhang, and Guo}]{pmlr-v220-kiseleva22a}
Julia Kiseleva, Alexey Skrynnik, Artem Zholus, Shrestha Mohanty, Negar Arabzadeh, Marc-Alexandre C\^{o}t\'e, Mohammad Aliannejadi, Milagro Teruel, Ziming Li, Mikhail Burtsev, Maartje ter Hoeve, Zoya Volovikova, Aleksandr Panov, Yuxuan Sun, Kavya Srinet, Arthur Szlam, Ahmed Awadallah, Seungeun Rho, Taehwan Kwon, Daniel Wontae~Nam, Felipe Bivort~Haiek, Edwin Zhang, Linar Abdrazakov, Guo Qingyam, Jason Zhang, and Zhibin Guo. 2022{\natexlab{b}}.
\newblock \href {https://proceedings.mlr.press/v220/kiseleva22a.html} {Interactive grounded language understanding in a collaborative environment: Retrospective on iglu 2022 competition}.
\newblock In \emph{Proceedings of the NeurIPS 2022 Competitions Track}, volume 220 of \emph{Proceedings of Machine Learning Research}, pages 204--216. PMLR.

\bibitem[{Kiseleva et~al.(2016{\natexlab{a}})Kiseleva, Williams, Hassan~Awadallah, Crook, Zitouni, and Anastasakos}]{kiseleva2016predicting}
Julia Kiseleva, Kyle Williams, Ahmed Hassan~Awadallah, Aidan~C Crook, Imed Zitouni, and Tasos Anastasakos. 2016{\natexlab{a}}.
\newblock Predicting user satisfaction with intelligent assistants.
\newblock In \emph{Proceedings of the 39th International ACM SIGIR conference on Research and Development in Information Retrieval}, pages 45--54.

\bibitem[{Kiseleva et~al.(2016{\natexlab{b}})Kiseleva, Williams, Jiang, Hassan~Awadallah, Crook, Zitouni, and Anastasakos}]{kiseleva2016understanding}
Julia Kiseleva, Kyle Williams, Jiepu Jiang, Ahmed Hassan~Awadallah, Aidan~C Crook, Imed Zitouni, and Tasos Anastasakos. 2016{\natexlab{b}}.
\newblock Understanding user satisfaction with intelligent assistants.
\newblock In \emph{Proceedings of the 2016 ACM on Conference on Human Information Interaction and Retrieval}, pages 121--130.

\bibitem[{K{\"o}pf et~al.(2024)K{\"o}pf, Kilcher, von R{\"u}tte, Anagnostidis, Tam, Stevens, Barhoum, Nguyen, Stanley, Nagyfi et~al.}]{kopf2024openassistant}
Andreas K{\"o}pf, Yannic Kilcher, Dimitri von R{\"u}tte, Sotiris Anagnostidis, Zhi~Rui Tam, Keith Stevens, Abdullah Barhoum, Duc Nguyen, Oliver Stanley, Rich{\'a}rd Nagyfi, et~al. 2024.
\newblock Openassistant conversations-democratizing large language model alignment.
\newblock \emph{Advances in Neural Information Processing Systems}, 36.

\bibitem[{Li et~al.(2023{\natexlab{a}})Li, Hammoud, Itani, Khizbullin, and Ghanem}]{li2023camel}
Guohao Li, Hasan Abed Al~Kader Hammoud, Hani Itani, Dmitrii Khizbullin, and Bernard Ghanem. 2023{\natexlab{a}}.
\newblock Camel: Communicative agents for" mind" exploration of large scale language model society.
\newblock \emph{arXiv preprint arXiv:2303.17760}.

\bibitem[{Li et~al.(2023{\natexlab{b}})Li, Cui, Kong, and Bi}]{li2023collaborative}
Qintong Li, Leyang Cui, Lingpeng Kong, and Wei Bi. 2023{\natexlab{b}}.
\newblock Collaborative evaluation: Exploring the synergy of large language models and humans for open-ended generation evaluation.
\newblock \emph{arXiv preprint arXiv:2310.19740}.

\bibitem[{Li et~al.(2020)Li, Kiseleva, Agarwal, de~Rijke, and White}]{li2020optimizing}
Ziming Li, Julia Kiseleva, Alekh Agarwal, Maarten de~Rijke, and Ryen~W White. 2020.
\newblock Optimizing interactive systems via data-driven objectives.
\newblock \emph{arXiv preprint arXiv:2006.12999}.

\bibitem[{Li et~al.(2021)Li, Park, Kiseleva, Kim, and Lee}]{li2021deus}
Ziming Li, Dookun Park, Julia Kiseleva, Young-Bum Kim, and Sungjin Lee. 2021.
\newblock Deus: A data-driven approach to estimate user satisfaction in multi-turn dialogues.
\newblock \emph{arXiv preprint arXiv:2103.01287}.

\bibitem[{Liang et~al.(2023{\natexlab{a}})Liang, Bommasani, Lee, Tsipras, Soylu, Yasunaga, Zhang, Narayanan, Wu, Kumar, Newman, Yuan, Yan, Zhang, Cosgrove, Manning, Ré, Acosta-Navas, Hudson, Zelikman, Durmus, Ladhak, Rong, Ren, Yao, Wang, Santhanam, Orr, Zheng, Yuksekgonul, Suzgun, Kim, Guha, Chatterji, Khattab, Henderson, Huang, Chi, Xie, Santurkar, Ganguli, Hashimoto, Icard, Zhang, Chaudhary, Wang, Li, Mai, Zhang, and Koreeda}]{liang2023holistic}
Percy Liang, Rishi Bommasani, Tony Lee, Dimitris Tsipras, Dilara Soylu, Michihiro Yasunaga, Yian Zhang, Deepak Narayanan, Yuhuai Wu, Ananya Kumar, Benjamin Newman, Binhang Yuan, Bobby Yan, Ce~Zhang, Christian Cosgrove, Christopher~D. Manning, Christopher Ré, Diana Acosta-Navas, Drew~A. Hudson, Eric Zelikman, Esin Durmus, Faisal Ladhak, Frieda Rong, Hongyu Ren, Huaxiu Yao, Jue Wang, Keshav Santhanam, Laurel Orr, Lucia Zheng, Mert Yuksekgonul, Mirac Suzgun, Nathan Kim, Neel Guha, Niladri Chatterji, Omar Khattab, Peter Henderson, Qian Huang, Ryan Chi, Sang~Michael Xie, Shibani Santurkar, Surya Ganguli, Tatsunori Hashimoto, Thomas Icard, Tianyi Zhang, Vishrav Chaudhary, William Wang, Xuechen Li, Yifan Mai, Yuhui Zhang, and Yuta Koreeda. 2023{\natexlab{a}}.
\newblock \href {http://arxiv.org/abs/2211.09110} {Holistic evaluation of language models}.

\bibitem[{Liang et~al.(2023{\natexlab{b}})Liang, He, Jiao, Wang, Wang, Wang, Yang, Tu, and Shi}]{liang2023encouraging}
Tian Liang, Zhiwei He, Wenxiang Jiao, Xing Wang, Yan Wang, Rui Wang, Yujiu Yang, Zhaopeng Tu, and Shuming Shi. 2023{\natexlab{b}}.
\newblock \href {http://arxiv.org/abs/2305.19118} {Encouraging divergent thinking in large language models through multi-agent debate}.

\bibitem[{Liu and Sun(2023)}]{liu2023voices}
Alex Liu and Min Sun. 2023.
\newblock From voices to validity: Leveraging large language models (llms) for textual analysis of policy stakeholder interviews.
\newblock \emph{arXiv preprint arXiv:2312.01202}.

\bibitem[{Liu et~al.(2024)Liu, Liu, Shi, Huang, Wang, Yang, and Zhang}]{liu2024exploring}
Fang Liu, Yang Liu, Lin Shi, Houkun Huang, Ruifeng Wang, Zhen Yang, and Li~Zhang. 2024.
\newblock Exploring and evaluating hallucinations in llm-powered code generation.
\newblock \emph{arXiv preprint arXiv:2404.00971}.

\bibitem[{Liu et~al.(2023)Liu, Yu, Zhang, Xu, Lei, Lai, Gu, Ding, Men, Yang et~al.}]{liu2023agentbench}
Xiao Liu, Hao Yu, Hanchen Zhang, Yifan Xu, Xuanyu Lei, Hanyu Lai, Yu~Gu, Hangliang Ding, Kaiwen Men, Kejuan Yang, et~al. 2023.
\newblock Agentbench: Evaluating llms as agents.
\newblock \emph{arXiv preprint arXiv:2308.03688}.

\bibitem[{Liu et~al.(2022)Liu, Schick, and Sch{\"u}tze}]{liu2022semantic}
Yanchen Liu, Timo Schick, and Hinrich Sch{\"u}tze. 2022.
\newblock Semantic-oriented unlabeled priming for large-scale language models.
\newblock \emph{arXiv preprint arXiv:2202.06133}.

\bibitem[{Mialon et~al.(2023)Mialon, Fourrier, Swift, Wolf, LeCun, and Scialom}]{mialon2023gaia}
Gr{\'e}goire Mialon, Cl{\'e}mentine Fourrier, Craig Swift, Thomas Wolf, Yann LeCun, and Thomas Scialom. 2023.
\newblock Gaia: a benchmark for general ai assistants.
\newblock \emph{arXiv preprint arXiv:2311.12983}.

\bibitem[{Myers et~al.(2023)Myers, He, Fang, Walke, Hansen-Estruch, Cheng, Jalobeanu, Kolobov, Dragan, and Levine}]{pmlr-v229-myers23a}
Vivek Myers, Andre~Wang He, Kuan Fang, Homer~Rich Walke, Philippe Hansen-Estruch, Ching-An Cheng, Mihai Jalobeanu, Andrey Kolobov, Anca Dragan, and Sergey Levine. 2023.
\newblock \href {https://proceedings.mlr.press/v229/myers23a.html} {Goal representations for instruction following: A semi-supervised language interface to control}.
\newblock In \emph{Proceedings of The 7th Conference on Robot Learning}, volume 229 of \emph{Proceedings of Machine Learning Research}, pages 3894--3908. PMLR.

\bibitem[{Pan et~al.(2024)Pan, Zhang, Tomlin, Zhou, Levine, and Suhr}]{pan2024autonomous}
Jiayi Pan, Yichi Zhang, Nicholas Tomlin, Yifei Zhou, Sergey Levine, and Alane Suhr. 2024.
\newblock Autonomous evaluation and refinement of digital agents.
\newblock \emph{arXiv preprint arXiv:2404.06474}.

\bibitem[{Reimers and Gurevych(2019)}]{reimers2019sentence}
Nils Reimers and Iryna Gurevych. 2019.
\newblock Sentence-bert: Sentence embeddings using siamese bert-networks.
\newblock \emph{arXiv preprint arXiv:1908.10084}.

\bibitem[{See et~al.(2019)See, Roller, Kiela, and Weston}]{see2019makes}
Abigail See, Stephen Roller, Douwe Kiela, and Jason Weston. 2019.
\newblock What makes a good conversation? how controllable attributes affect human judgments.
\newblock \emph{arXiv preprint arXiv:1902.08654}.

\bibitem[{Sepliarskaia et~al.(2018)Sepliarskaia, Kiseleva, Radlinski, and de~Rijke}]{sepliarskaia2018preference}
Anna Sepliarskaia, Julia Kiseleva, Filip Radlinski, and Maarten de~Rijke. 2018.
\newblock Preference elicitation as an optimization problem.
\newblock In \emph{Proceedings of the 12th ACM Conference on Recommender Systems}, pages 172--180.

\bibitem[{Shridhar et~al.(2019)Shridhar, Thomason, Gordon, Bisk, Han, Mottaghi, Zettlemoyer, and Fox}]{DBLP:journals/corr/abs-1912-01734}
Mohit Shridhar, Jesse Thomason, Daniel Gordon, Yonatan Bisk, Winson Han, Roozbeh Mottaghi, Luke Zettlemoyer, and Dieter Fox. 2019.
\newblock \href {http://arxiv.org/abs/1912.01734} {{ALFRED:} {A} benchmark for interpreting grounded instructions for everyday tasks}.
\newblock \emph{CoRR}, abs/1912.01734.

\bibitem[{Shridhar et~al.(2020{\natexlab{a}})Shridhar, Thomason, Gordon, Bisk, Han, Mottaghi, Zettlemoyer, and Fox}]{shridhar2020alfred}
Mohit Shridhar, Jesse Thomason, Daniel Gordon, Yonatan Bisk, Winson Han, Roozbeh Mottaghi, Luke Zettlemoyer, and Dieter Fox. 2020{\natexlab{a}}.
\newblock Alfred: A benchmark for interpreting grounded instructions for everyday tasks.
\newblock In \emph{Proceedings of the IEEE/CVF conference on computer vision and pattern recognition}, pages 10740--10749.

\bibitem[{Shridhar et~al.(2020{\natexlab{b}})Shridhar, Yuan, C{\^o}t{\'e}, Bisk, Trischler, and Hausknecht}]{shridhar2020alfworld}
Mohit Shridhar, Xingdi Yuan, Marc-Alexandre C{\^o}t{\'e}, Yonatan Bisk, Adam Trischler, and Matthew Hausknecht. 2020{\natexlab{b}}.
\newblock Alfworld: Aligning text and embodied environments for interactive learning.
\newblock \emph{arXiv preprint arXiv:2010.03768}.

\bibitem[{Stahl and Eke(2024)}]{stahl2024ethics}
Bernd~Carsten Stahl and Damian Eke. 2024.
\newblock The ethics of chatgpt--exploring the ethical issues of an emerging technology.
\newblock \emph{International Journal of Information Management}, 74:102700.

\bibitem[{Talebirad and Nadiri(2023)}]{talebirad2023multi}
Yashar Talebirad and Amirhossein Nadiri. 2023.
\newblock Multi-agent collaboration: Harnessing the power of intelligent llm agents.
\newblock \emph{arXiv preprint arXiv:2306.03314}.

\bibitem[{Tjuatja et~al.(2023)Tjuatja, Chen, Wu, Talwalkar, and Neubig}]{tjuatja2023llms}
Lindia Tjuatja, Valerie Chen, Sherry~Tongshuang Wu, Ameet Talwalkar, and Graham Neubig. 2023.
\newblock Do llms exhibit human-like response biases? a case study in survey design.
\newblock \emph{arXiv preprint arXiv:2311.04076}.

\bibitem[{Vahtola et~al.(2022)Vahtola, Creutz, and Tiedemann}]{vahtola2022not}
Teemu Vahtola, Mathias Creutz, and J{\"o}rg Tiedemann. 2022.
\newblock It is not easy to detect paraphrases: Analysing semantic similarity with antonyms and negation using the new semantoneg benchmark.
\newblock In \emph{Proceedings of the Fifth BlackboxNLP Workshop on Analyzing and Interpreting Neural Networks for NLP}, pages 249--262.

\bibitem[{Wang et~al.(2023{\natexlab{a}})Wang, Liang, Meng, Sun, Shi, Li, Xu, Qu, and Zhou}]{wang2023chatgpt}
Jiaan Wang, Yunlong Liang, Fandong Meng, Zengkui Sun, Haoxiang Shi, Zhixu Li, Jinan Xu, Jianfeng Qu, and Jie Zhou. 2023{\natexlab{a}}.
\newblock Is chatgpt a good nlg evaluator? a preliminary study.
\newblock In \emph{Proceedings of the 4th New Frontiers in Summarization Workshop}, pages 1--11.

\bibitem[{Wang et~al.(2023{\natexlab{b}})Wang, HU, Hou, Chen, Zheng, Wang, Yang, Ye, Huang, Geng, Jiao, Zhang, and Xie}]{wang2023robustness}
Jindong Wang, Xixu HU, Wenxin Hou, Hao Chen, Runkai Zheng, Yidong Wang, Linyi Yang, Wei Ye, Haojun Huang, Xiubo Geng, Binxing Jiao, Yue Zhang, and Xing Xie. 2023{\natexlab{b}}.
\newblock \href {https://openreview.net/forum?id=uw6HSkgoM29} {On the robustness of chat{GPT}: An adversarial and out-of-distribution perspective}.
\newblock In \emph{ICLR 2023 Workshop on Trustworthy and Reliable Large-Scale Machine Learning Models}.

\bibitem[{Wang et~al.(2023{\natexlab{c}})Wang, Li, Chen, Zhu, Lin, Cao, Liu, Liu, and Sui}]{wang2023large}
Peiyi Wang, Lei Li, Liang Chen, Dawei Zhu, Binghuai Lin, Yunbo Cao, Qi~Liu, Tianyu Liu, and Zhifang Sui. 2023{\natexlab{c}}.
\newblock Large language models are not fair evaluators.
\newblock \emph{arXiv preprint arXiv:2305.17926}.

\bibitem[{Williams et~al.(2016{\natexlab{a}})Williams, Kiseleva, Crook, Zitouni, Awadallah, and Khabsa}]{williams2016detecting}
Kyle Williams, Julia Kiseleva, Aidan~C Crook, Imed Zitouni, Ahmed~Hassan Awadallah, and Madian Khabsa. 2016{\natexlab{a}}.
\newblock Detecting good abandonment in mobile search.
\newblock In \emph{Proceedings of the 25th International Conference on World Wide Web}, pages 495--505.

\bibitem[{Williams et~al.(2016{\natexlab{b}})Williams, Kiseleva, Crook, Zitouni, Awadallah, and Khabsa}]{williams2016your}
Kyle Williams, Julia Kiseleva, Aidan~C Crook, Imed Zitouni, Ahmed~Hassan Awadallah, and Madian Khabsa. 2016{\natexlab{b}}.
\newblock Is this your final answer? evaluating the effect of answers on good abandonment in mobile search.
\newblock In \emph{Proceedings of the 39th International ACM SIGIR conference on Research and Development in Information Retrieval}, pages 889--892.

\bibitem[{Williams and Zitouni(2017)}]{williams2017does}
Kyle Williams and Imed Zitouni. 2017.
\newblock Does that mean you're happy? rnn-based modeling of user interaction sequences to detect good abandonment.
\newblock In \emph{Proceedings of the 2017 ACM on Conference on Information and Knowledge Management}, pages 727--736.

\bibitem[{Winograd(1972)}]{winograd1972understanding}
Terry Winograd. 1972.
\newblock Understanding natural language.
\newblock \emph{Cognitive psychology}, 3(1):1--191.

\bibitem[{Wu et~al.(2023)Wu, Bansal, Zhang, Wu, Zhang, Zhu, Li, Jiang, Zhang, and Wang}]{wu2023autogen}
Qingyun Wu, Gagan Bansal, Jieyu Zhang, Yiran Wu, Shaokun Zhang, Erkang Zhu, Beibin Li, Li~Jiang, Xiaoyun Zhang, and Chi Wang. 2023.
\newblock Autogen: Enabling next-gen llm applications via multi-agent conversation framework.
\newblock \emph{arXiv preprint arXiv:2308.08155}.

\bibitem[{Yao et~al.(2022)Yao, Zhao, Yu, Du, Shafran, Narasimhan, and Cao}]{yao2022react}
Shunyu Yao, Jeffrey Zhao, Dian Yu, Nan Du, Izhak Shafran, Karthik Narasimhan, and Yuan Cao. 2022.
\newblock React: Synergizing reasoning and acting in language models.
\newblock \emph{arXiv preprint arXiv:2210.03629}.

\bibitem[{Zhang et~al.(2023)Zhang, Cai, Fu, Yuan, and Lu}]{zhang2023creative}
Chi Zhang, Penglin Cai, Yuhui Fu, Haoqi Yuan, and Zongqing Lu. 2023.
\newblock Creative agents: Empowering agents with imagination for creative tasks.
\newblock \emph{arXiv preprint arXiv:2312.02519}.

\bibitem[{Ziyu et~al.(2023)Ziyu, Qiguang, Longxuan, Mingda, Yi, Yushan, Haopeng, Weinan, and Liu}]{ziyu-etal-2023-lens}
Zhuang Ziyu, Chen Qiguang, Ma~Longxuan, Li~Mingda, Han Yi, Qian Yushan, Bai Haopeng, Zhang Weinan, and Ting Liu. 2023.
\newblock \href {https://aclanthology.org/2023.ccl-2.8} {Through the lens of core competency: Survey on evaluation of large language models}.
\newblock In \emph{Proceedings of the 22nd Chinese National Conference on Computational Linguistics (Volume 2: Frontier Forum)}, pages 88--109, Harbin, China. Chinese Information Processing Society of China.

\end{thebibliography}
\bibliographystyle{acl_natbib}

\appendix
\newpage
\clearpage

\section{Appendix}
\label{sec:appendix}

\subsection{Task Examples}
\label{sec:task-example}
In \fig~\ref{fig:math-example} and \ref{alf-example}, we display examples of Math problems and ALFWorld house-holding tasks solved with AutoGen.

\begin{figure*}[h!]
\begin{tcolorbox}[colback=gray!5!white,colframe=gray!75!black]
{
  "problem": "How many positive and negative integers is $12$ a multiple of?"
  
  "level": "Level 5"
  
  "type": "Prealgebra"

    "response\_with\_ans": 
    
  "messages": [
  
    \{     
    
    "content": "How many positive and negative integers is $12$ a multiple of?",
      
      "role": "user"
    
    \},
    
    \{
    
    "content": "To determine how many positive and negative integers 12 is a multiple of, we need to find the divisors of 12. Then we will double the number of divisors, as each positive divisor has a corresponding negative divisor.
      Let's use a Python code snippet to find all the divisors of 12 and count them.",
      \begin{lstlisting}[language=Python]
num = 12
divisors = [i for i in range(1, num + 1) if num \% i == 0]
positive_divisors_count = len(divisors)
negative_divisors_count = positive_divisors_count
total_divisors_count = positive_divisors_count + negative_divisors_count
print(f"Positive divisors of 12: {divisors}")
divisors_negative = [-x for x in divisors]
print(f"Negative divisors of 12: {divisors_negative}")
print(f"Total divisors count: {total_divisors_count}")
\end{lstlisting}

    Execute this code snippet to find the divisors of 12 and count how many positive and negative integers 12 is a multiple of.
    
      "role": "assistant"
  
    \}
 
  ],
  
  "time": 17.360238790512085

}
 
\end{tcolorbox}
\caption{An example of Math Problem solving sample with solution generated by AutoGen}
\label{fig:math-example}
\end{figure*}

\begin{figure*}[]
\begin{Verbatim}[fontsize=\footnotesize, breaklines]
{
    { "content": "Perform actions and interact with a household to solve a task. At the beginning of your interactions, you will be given the detailed description of the current environment and your goal to accomplish. For each of your turn, you should choose from two actions: "think" or     "Action". If you choose "think", you should first think about the current condition and plan for your future actions, and then output your action in this turn. Your output must strictly follow this format:"think: your thoughts."
    "Action: your next action\\n"; If you choose "ACTION", you should directly output the action in this turn. Your output must strictly follow this format:"ACTION: your next action".
    After each of your turn, the environment will give you immediate feedback based on which you should plan your next few steps. if the envrionment output "Nothing happened", that means the previous action is invalid and you should try more options.
    Reminder: 
    1. the action must be chosen from the given available actions. Any actions except provided available actions will be regarded as illegal.
    2. Take the target household immediatly after you find it.
    3. Reply 'TERMINATE' only under two circumstances: a). The task has given you clear instructions to return 'TERMINATE' b). The task seems insolvable.\\n Here are two examples.\nYou are in the middle of a room. Looking quickly around you, you see a cabinet 13, a cabinet 12, a cabinet 11, a cabinet 10, a cabinet 9, a cabinet 8, a cabinet 7, a cabinet 6, a cabinet 5, a cabinet 4, a cabinet 3, a cabinet 2, a cabinet 1, a coffeemachine 1, a countertop 1, a diningtable 1, a drawer 1, a fridge 1, a garbagecan 1, a microwave 1, a shelf 3, a shelf 2, a shelf 1, a sinkbasin 1, a stoveburner 4, a stoveburner 3, a stoveburner 2, a stoveburner 1, and a toaster 1.\nYour task is to: put a hot apple in fridge.\n",
        "role": "user"},
    {   "content": "THOUGHT: The task is to put a hot apple in fridge. So i need to first find an apple and heat it. First I need to find an apple. An apple is more likely to appear in diningtable (1), coffeetable (1), fridge (1), drawer (1), cabinet (1-13), garbagecan (1). I can check one by one, starting with fridge 1.\n ACTION: go to diningtable 1\n",
        "role": "assistant"},
    {   "content": "On the diningtable 1, you see a apple 1, a bread 1, a fork 3, a fork 2, a fork 1, a mug 2, a peppershaker 3, a plate 2, a pot 1, a soapbottle 3, a spatula 1, a spoon 3, a spoon 2, a spoon 1, a winebottle 3, and a winebottle 2.\n",
        "role": "user"},
    {   "content": "ACTION: take apple 1 from diningtable 1\n",
        "role": "assistant"},
    {   "content": "You pick up the apple 1 from the diningtable 1.\n",
        "role": "user"},
    {   "content": "THOUGHT: I noticed that there is a microwave 1 in the room, i will use it to heat the apple.\n ACTION: go to microwave 1\n",
        "role": "assistant"},
    ...
    ...
 
    {   "content": "You open the microwave 1. The microwave 1 is open. In it, you see a egg 1.",
        "role": "user"
    },
    {   "content": "ACTION: heat mug 2 with microwave 1",
        "role": "assistant"},
    {    "content": "You heat the mug 2 using the microwave 1.Task success, now reply TERMINATE\n",
        "role": "user"},
    {   "content": "TERMINATE",
        "role": "assistant"},
    {   "content": "TERMINATE",
        "role": "user"}
}
\end{Verbatim}
\caption{An example of ALFWorld Householding task from ``Pick, Heat, then Place'' category.}

\label{alf-example}

\end{figure*}

\subsection{Math Problem Solving Benchmark}
\label{apd:math}
For math problem solving, although success is clearly defined, multiple solutions exist for accomplishing the objectives.
The MATH dataset, originally is a substantial collection of 12,500 challenging mathematics problems from high school competitions~\cite{hendrycks2021measuring}. Each problem comes with a step-by-step solution, enabling models to learn how to generate both derivations and explanations. The dataset covers a wide range of mathematical subjects and is tagged by difficulty levels, offering a nuanced measure of model performance across various aspects of mathematical problem-solving.

This dataset is particularly suitable for testing multi-agent systems for several reason including: 
\begin{enumerate*}[label=(\roman*)]
\item The problems in the MATH dataset are not simple computations but require a deep understanding of mathematical concepts, heuristics, and problem-solving strategies.  
\item Since the dataset includes step-by-step solutions, it allows for the assessment of an agent's ability to learn and reason through a problem, not just its ability to arrive at the correct answer. 
\item The variety of subjects and difficulty levels in the MATH dataset enables a comprehensive evaluation of a system’s versatility and adaptability in different mathematical domains which  is crucial for multi-agent systems that are expected to operate across a range of scenarios.
\end{enumerate*}

Similar to math problem experimental setup in \citet{wu2023autogen}, we carry out two experimental evaluations which involves 120 problems from the most challenging category, and includes 20 problems each from six different categories, of number theory, counting and probability, prealgebra, algebra, intermediate algebra, and precalculus.

\begin{table}[t]
    \centering
        \caption{Verification Criteria for ALFWorld Housholding Tasks. }

    \scalebox{0.56}{
    \begin{tabular}{>{\raggedright\arraybackslash}p{1.9cm}>{\raggedright\arraybackslash}p{6.3cm}>{\raggedright\arraybackslash}p{4.5cm}}
        \hline \hline
        \textbf{Criteria} & \textbf{Description} & \begin{center}
        \vspace{-1em}
            \textbf{Accepted Values}
                    \vspace{-1em}
        \end{center}  \\
        \hline
        Task Understanding & How well the participant was able to comprehend the problem set and follow the task instructions  & 
                              \vspace{-0.5em}
\begin{enumerate}[label=--,itemsep=0em,parsep=-0.2em,topsep=-4em]
            \item Excellent (4)
            \item  Good (3)
            \item Average (2)
            \item Poor (1)
            \item Terrible (0)
          \end{enumerate} 
        \\[-1em]
        \hline
       Plan Making & The ability of the participant to strategize and make a plan for tackling the task. & 
                                       \vspace{-0.5em}
 \begin{enumerate}[label=--,itemsep=0em,parsep=-0.2em,topsep=-4em]
            \item Excellent (4)
            \item  Good (3)
            \item Average (2)
            \item Poor (1)
            \item Terrible (0)
        \end{enumerate} 
       \\[-1em]
        \hline
        Action Decision & The participant's decision-making skills in choosing the right action to perform. & 
                                       \vspace{-0.5em}
 \begin{enumerate}[label=--,itemsep=0em,parsep=-0.2em,topsep=-4em]
            \item Excellent (4)
            \item  Good (3)
            \item Average (2)
            \item Poor (1)
            \item Terrible (0)
 
        \end{enumerate}
    \\[-1em]
        \hline
        Action Execution & How effectively the participant is able to execute the chosen action. &              
                                       \vspace{-0.5em}
 \begin{enumerate}[label=--,itemsep=0em,parsep=-0.2em,topsep=-4em]
            \item Excellent (4)
            \item  Good (3)
            \item Average (2)
            \item Poor (1)
            \item Terrible (0)
 
        \end{enumerate}
    \\[-1em]
        \hline
        Response to Feedback & How well the participant adapts his/her next steps based on the feedback from the environment & 
                                       \vspace{-0.5em}
 \begin{enumerate}[label=--,itemsep=0em,parsep=-0.1em,topsep=-4em]
            \item Excellent (4)
            \item Good (3)
            \item Average (2)
            \item Poor (1)
            \item Terrible (0)
        \end{enumerate}
        \\[-1em]
        \hline
        Correctness of Action &The correctness of the action performed by the participant with respect to the available actions and the current context & 
        \vspace{-0.5em}
  \begin{enumerate}[label=--,itemsep=0em,parsep=-0.1em,topsep=-4em]
            \item Correct  (1)
            \item Incorrect (0)
        \end{enumerate}
        \\
        \hline
                Use of Terminate & Whether the participant uses the 'TERMINATE' command appropriately & 
                \vspace{-0.5em}
  \begin{enumerate}[label=--,itemsep=0em,parsep=-0.1em,topsep=-4em]
            \item Appropriate  (1)
            \item Inappropriate (0)
        \end{enumerate}
        \\[-1em]
         \hline \hline
    \end{tabular}
    }
    \label{tab:criteria-alf}
\end{table}

\subsection{ALFWorld House-holding Task}
\label{apd:alf}

\begin{figure*}[t]
\vspace{-2em}
\centering
  \includegraphics[clip, trim=0cm 0cm 0cm 0cm,scale=0.65]{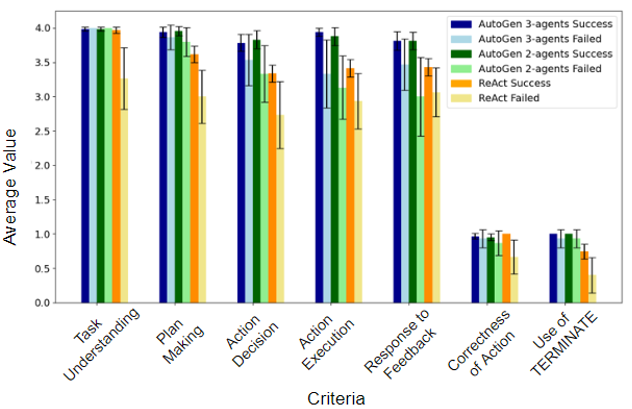}
  \caption{AgentEval assessment of three different solutions on ALFWorld house-holding Tasks categorized by success and failed cases.}
\label{img:alf-exp1}
\end{figure*}

\subsubsection{ALFWorld Dataset}
\label{apd:alf-data}

ALFWorld, presents a set of language-based interactive decision-making tasks within simulated household environments \cite{shridhar2020alfworld}. This benchmark is distinguished by its diversity of tasks, offering a comprehensive platform for testing AI and multi-agent systems. This benchmark is particularly suited for such evaluations because first, ALFWorld is the first interactive parallel environment that aligns text descriptions and commands with physically embodied robotic simulation. It extends two prior works: TextWorld, an engine for interactive text-based games, and ALFRED, a large-scale dataset for vision-language instruction following in embodied environments~\cite{shridhar2020alfred,cote2019textworld}.
The cross-modality framework of this benchmark allows for a variety of embodied tasks with corresponding text-based counterparts, allowing agents to be trained and evaluated in both the language and embodied world. In addition, ALFWorld supports the development of agents that can reason both abstractly and execute actions concretely, mimicking human-like decision-making processes in varying contexts.
Finally, the dataset's inclusion of a wide range of tasks, from household chores to more intricate problem-solving scenarios, provides a comprehensive testbed for evaluating the adaptability and problem-solving capabilities of AI and multi-agent systems. In general, the dataset allows agents to explore, interact, and learn in an abstract language environment before dealing with the complexities of an embodied environment.

\subsubsection{Solutions for ALFWorld benchmark}
\label{apd:alf-sol}
As for the solutions to assess for ALFWorld Household tasks, similar to \cite{wu2023autogen}, we consider ReAct~\cite{yao2022react} as well as AutoGen with two agents and AutoGen with three agents \cite{wu2023autogen}. ReAct is an agent that operates within the ALFWorld environments and is responsible for suggesting plans and executing actions. On the other hand, AutoGen Two-Agent System consists of an LLM-backed assistant agent responsible for suggesting plans, and an executor agent responsible for executing actions in the ALFWorld environments. Both ReAct and this solution occasionally struggles with leveraging basic commonsense knowledge about the physical world, which can lead to repetitive errors and getting stuck in loops.In AutoGen with three agents, a grounding agent is provided just for the sake of critical common sense knowledge whenever the system exhibits early signs of recurring errors.

\subsubsection{AgentEval Results for ALFWorld}
\label{apd:alf-main-res}

To study the generalizability of AgentEval, we repeat the experiments in \ref{sec:agenteval-math} for AlfWorld, in which real-world household environments are emulated through textual interfaces \cite{shridhar2020alfworld}. We provide the criteria created for this task as well as the results for three solutions of this task in Tab. \ref{tab:criteria-alf} and \fig~\ref{img:alf-exp1}, respectively.
Following the extraction of a set of criteria as detailed in Tab. \ref{tab:criteria-alf}, these criteria are passed to the \qa for quantification on each sample. Figure \ref{img:alf-exp1} presents the results for three introduced solutions: AutoGen with two agents, AutoGen with three agents, and ReAct, on the 134-test set from \cite{wu2023autogen}. Further, we also verify the measured performance of AgentEval, by comparing the failed/success cases in the right plot of Figure \ref{img:alf-exp1}. As confirmed here and in math problem solving, AgentEval is clearly able to distinguish between different solutions in terms of different criterion. 

We provide an example of AgentEval applied to the ALFWorld house-holding task, as mentioned in Sec.~\ref{sec:agenteval-math} in which real-world household environments is emulated through textual interfaces \cite{shridhar2020alfworld}. When running the \ca on this task, it identified specific criteria such as ``Task understanding'', ``Plan making'' and ``Response to Feedback'' as outlined in Tab.~\ref{tab:criteria-alf}. We consulted researchers deeply involved with these tasks, and their expertise confirmed that these criteria are critically relevant and significant similar to~\cite{li2023collaborative}.
For example, given that these tasks are language-based and require interactive decision-making, an agent in ALFWorld is tasked with high-level objectives, such as placing a hot apple in the fridge, and must navigate and interact with a simulated household environment to achieve these objectives. Therefore, criteria displayed in Tab.~\ref{tab:criteria-alf} satisfy the assessment of this task. While the criteria are pretty self-descriptive, about the criterion ``Use of TERMINATE'' we note that the agent is prompted to use the term ``TERMINATE'' upon task completion, which is closely correlated with task success.

\begin{figure*}[t]
\centering
\vspace{-2em}
\includegraphics[width=1.02\textwidth]{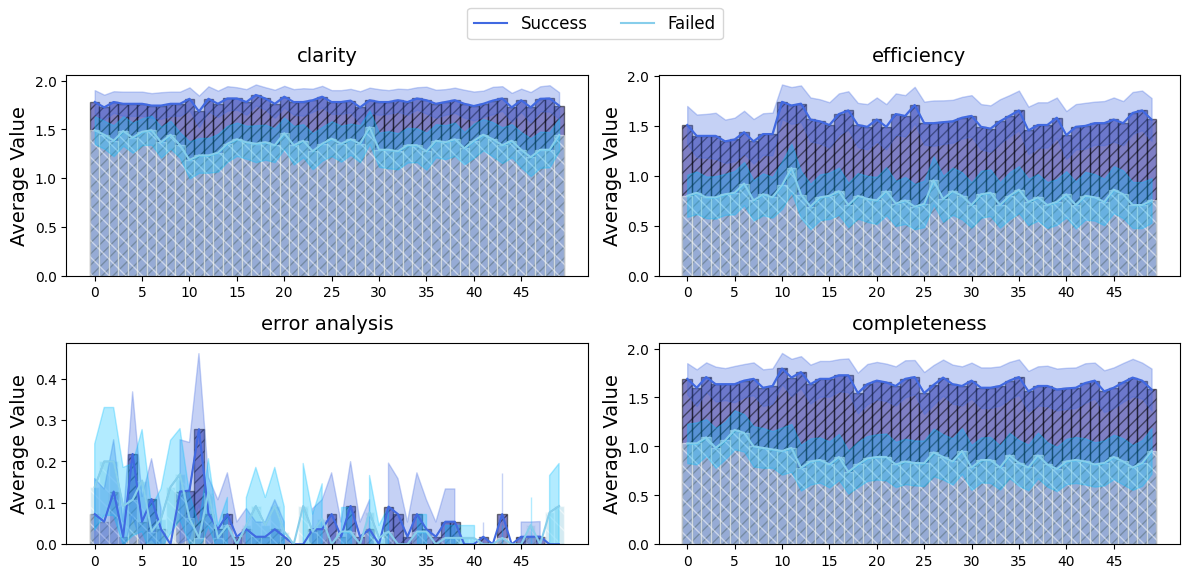}
\caption{Quantifier Robustness on criteria of Math Problem Solving problem. Each bar represent the average performance of success (dark blue "//") and failed (light blue ``\textbackslash\textbackslash'') cases and 95\% interval on each set is  shaded across the average point.  The two plots are overlaid.}
\label{fig:q-robust-first}
\end{figure*}

Following the extraction of a set of criteria as detailed in Tab \ref{tab:criteria-alf}, these criteria are passed to the \qa for quantification on each sample. Figure \ref{img:alf-exp1} presents the results for three introduced solutions: AutoGen with 2 agents, AutoGen with 3 agents, and ReAct, on the 134-test set from \cite{wu2023autogen}.
It is important to note that all criteria, except ``Use of TERMINATE'' and ``Correctness of Action'' employ a five-level grading system, while these two criteria are binary. From this figure, it is evident that ReACT performs notably worse across all criteria, while AutoGen with 2 agents and 3 agents demonstrate competitive performance. 
We also categorizes the 134 games into  groups of  failed and successful ones. Similar to \fig~\ref{img:math-exp1}, darker colors represent performance in successful cases for each solution, while lighter colors represent performance in failed cases. AutoGen 3-agent, AutoGen 2-agent, and ReAct are represented by blue, green, and orange, respectively. For most criteria, the distinction between failed and successful cases is clear, even within a 95\% confidence interval. However, for certain criteria, such as ``Task understanding'' all solutions, whether they failed or succeeded, exhibit very similar performance. This could be interpreted as either (1) all solutions have a good understanding of the task, even if they fail to complete it, (2) this criterion may be redundant, as it does not provide additional information among these three solutions or (3) the \qa is unable to score the criterion in a meaningful way. We refrain from concluding which criteria are most suitable for this specific task. Instead, we emphasize the importance of conducting a more in-depth analysis of performance beyond success rates, tailored to one's goals and application requirements. Later, we show that how using \va could be helpful in identifying criteria with higher discriminative power and more robustness.

\subsection{Robustness Analysis}

\subsubsection{Similar Criteria}
As explained in Section \ref{sec:c-robust}, there might be cases where some criteria are pointing to the same concepts with different wordings. In these cases, we need to merge the similar criteria to avoid having redundant criteria.
Table \ref{tab:sim_criteria} shows some of these examples.

\begin{table}[t]
    \centering
    \caption{Example pairs of similar criteria.}
    \scalebox{0.72}{
    \begin{tabular}{p{10cm}}
    \hline\hline
      - Problem Difficulty: The complexity of the math problem that has been solved. 
 \\   - Problem Complexity: The level of difficulty of the problem.\\ \hline
      - Innovativeness: The novelty and creativity in the approach to solve the problem 	 
   \\ - Innovation: The ability to solve a problem using a unique or creative method not commonly known.\\ \hline
      - Time Taken: The time taken to solve the problem. 	 
      \\ - Time to Completion: The amount of time taken to solve the problem completely \\\hline
       - Understandability: The clarity and ease of comprehension of the solution provided. 	
    \\ - Readability: How easy it is to comprehend the provided solution. \\\hline\hline
    \end{tabular}}
    \label{tab:sim_criteria}
\end{table}

\subsubsection{Quantifier Robustness}

\label{apd:robustness} 
To study the robustness of the \qa, we selected a specific subset of criteria related to mathematical problems, as detailed in Table \ref{tab:criteria-math}, and conducted 50 runs of the quantifier agent on the 120 problems described in Section \ref{sec:math-problems}. Our expectation is to observe consistent quantified performance for each of the criteria. In \fig~ \ref{fig:q-robust-first}, we present the distribution of quantified performance across 50 runs for both successful and failed cases, focusing on the five selected criteria. A consistently horizontal performance trend indicates greater robustness in the quantifier, whereas more fluctuations in the figure suggest less robustness and a noisier performance of the agent.

As shown in the results, for four out of the five generated criteria, we consistently observe steady performance. Not only do the success cases consistently outperform the failed cases, but their performance also falls within a similar range across runs. However, when it comes to the ``error analysis'' criterion, we observe a more variable performance of the quantifier. It does not consistently predict one group (success or failed) to perform better than the other, and the quantifier's performance varies across different runs. This suggests that the AgentEval tool may not exhibit promising robustness for this particular criterion. The underlying issues could be either the criterion itself lacks clarity and appropriateness for the task, or the \qa struggles to quantify this criterion effectively. In either case, it is advisable to either modify or eliminate this criterion to enhance trustworthiness and reliability. We further show that \va is designed to take care of such criteria.

We recognize the importance of thoroughly investigating the robustness of each criterion in quantification studies. This analysis is crucial as it sheds light on the stability of each criterion. Moreover, when ground truths are available, such as in cases of success versus failure, they provide a benchmark to validate our assessments. Additionally, it is important to acknowledge that not all criteria exhibit the same level of robustness. This variability demands careful consideration during evaluations, especially given the non-deterministic nature of LLMs. Such awareness is essential to ensure the reliability and accuracy of our assessments in the dynamic field of LLMs.

\subsection{\va}
Algorithm \ref{alg:verifier} shows how \va works.
To make \va works, we need to study the stability of proposed criteria as well as how robust they are w.r.t the injected noise.

\begin{figure*}[h!]
\centering
\vspace{-15em}
  \includegraphics[clip, trim=0cm 0cm 0cm 0cm,scale=0.4]{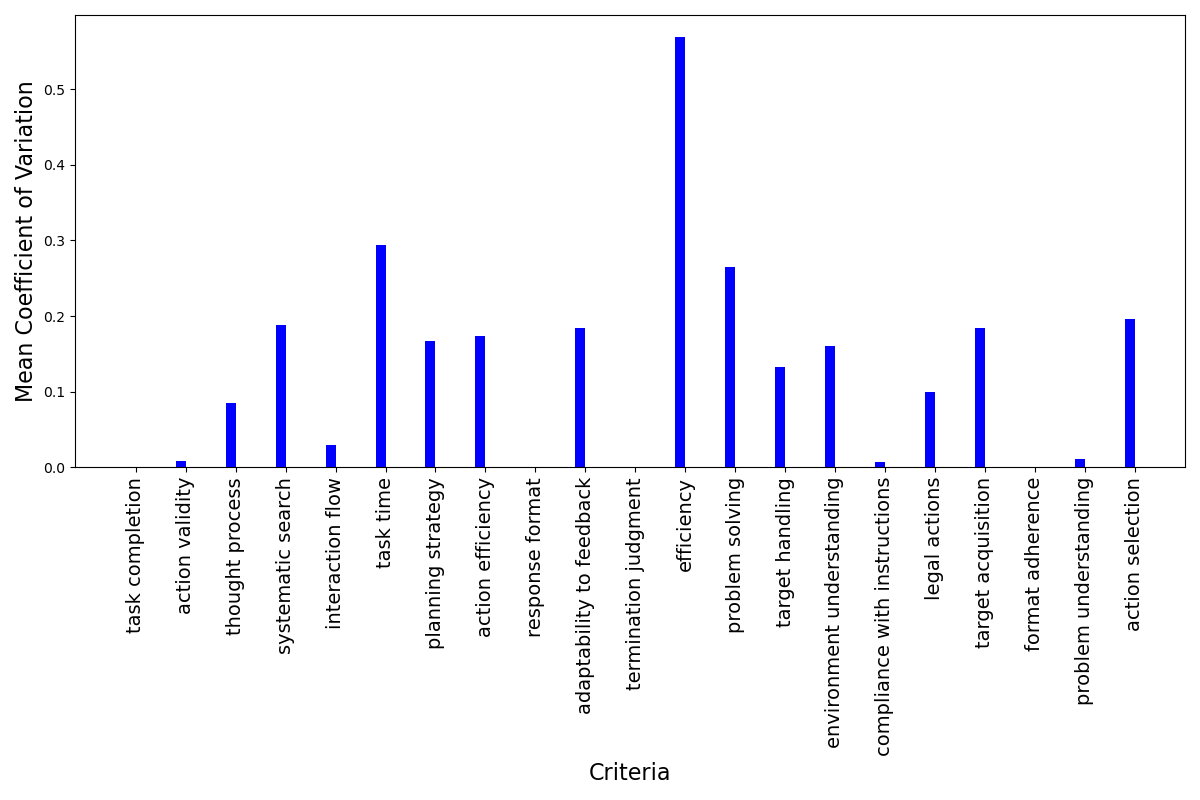}
\caption{Evaluating the \qa's robustness on ALFWorld dataset: the mean coefficient of variation of quantified results across $n=18$ seeds.}
\label{fig:ALFWorld_mean_coeff_var_comp}
\end{figure*}

\subsubsection{Criteria Robustness}

we first report the full criteria list for Math problems solving and ALFWorld household tasks when running the \ca and \qa for 50 times after consolidation (as described in section \ref{sec:c-robust}) in Tab \ref{tab:full-criteria-math} and \ref{tab:full-criteria-alf}. 
This process would exclude criteria that have mean standard deviation above a certain threshold and criteria that have a higher or equivalent average score for adversarial task output than the original task output. This does not necessarily mean these criteria are bad criteria, but rather suggests the \qa may not be able to reliably quantify these criteria and thus it might be better to exclude them from the final score assigned to a sample.
As such, similar to \fig \ref{fig:num_seeds}, we report the mean of coefficient variation for ALFWorld task in \fig \ref{fig:ALFWorld_mean_coeff_var_comp}. We note that having almost all of the coefficient below 0.5 indicate high level of robustness of \qa on the verified set of criteria by \va on AlfWorld dataset. 

\begin{figure}[htb]
  \centering
  \begin{minipage}{\linewidth}
        \vspace{-1em}
    \begin{algorithm}[H]  
      \caption{ \va } 
      \label{alg:verifier}
      \begin{algorithmic}[1]
        \For {$i=1,2,\ldots,50$}
          \State Run \emph{CriticAgent} with $seed=i$ to obtain a set of criteria $C_i$
        \EndFor
        \State Obtain \emph{summarized\_criteria} by using another LLM agent to summarize $C_1,C_2,\ldots,C_{50}$. 
        \For {$i=1,2,\ldots,18$}
          \ForAll{$s$ in $S$}
            \State Run \emph{QuantifierAgent} with $seed=i$ on sample $s$
          \EndFor 
        \EndFor
        \ForAll{\emph{crit} in \emph{summarized\_criteria}}
          \ForAll{$s$ in $S$}
            \State Compute the \emph{coefficient of variation} of $s$'s quantified result with respect to \emph{crit} across all $seed$
          \EndFor 
          \State Compute \emph{mean coefficient of variation} by averaging all sample's coefficient of variation
        \EndFor 

        \State \emph{final\_criteria} $\gets []$
        \ForAll{\emph{crit} in \emph{summarized\_criteria}}
          \If{\emph{crit} has a \emph{mean coefficient of variation} within a certain range, and \emph{crit} has decent adversarial testing performance}
            \State Add \emph{crit} to \emph{final\_criteria}
          \EndIf
        \EndFor

        \State To evaluate future tasks, use \emph{final\_criteria} with \emph{QuantifierAgent}.
      \end{algorithmic}
    \end{algorithm}
    \vspace{-2em}
  \end{minipage}
\end{figure}

\subsubsection{Adversarial Attacks}

\begin{figure*}[h!]
\centering
\vspace{-25em}
  \includegraphics[clip, trim=0cm 0cm 0cm 0cm,scale=0.4]{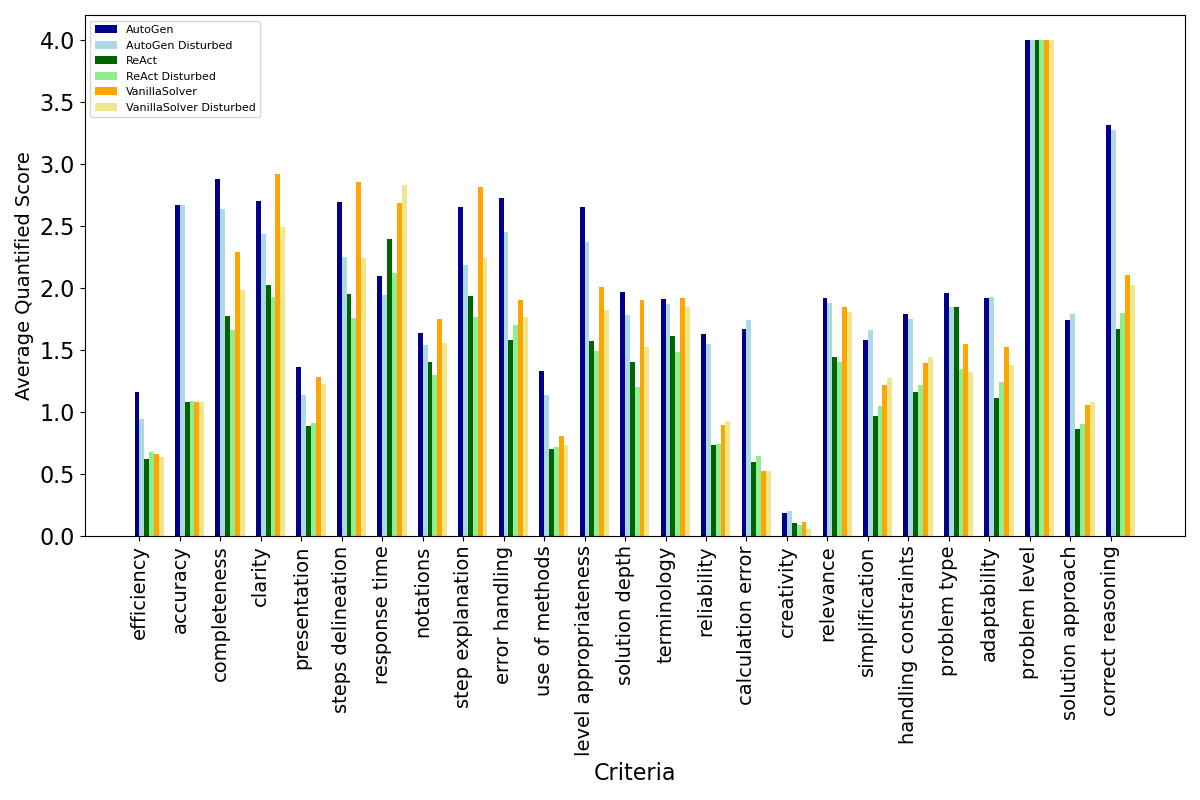}
\caption{\qa Verification on original set of task solutions against the disturbed task solutions on Math Problem Solving dataset.}
\label{fig:adversarial_results}
\end{figure*}

We construct adversarial samples by randomly dropping a portion of sentences in the LLM assistant's response from the original task output. We verify the \qa against the adversarial samples. We used three different benchmarks for adversarial testing, namely  AutoGen, ReAct and Vanilla Solver. As shown in \fig~\ref{fig:adversarial_results} for the ALFWorld dataset), in most cases the \qa quantifies the adversarial task output to be worse off than the original task output. We believe that such an analysis of the quantifier agent’s performance will yield valuable insights for enhancing reliability, trustworthiness, and explainability in performance evaluation.

One interesting observation here is that there maybe interdependence among some criteria. For example \emph{level\:appropriatness} is defined as \emph{"How well-suited the solution provided by the system is for the given problem's level"}, which is dependent on the criterion \emph{problem\:level}. This observation gives insight into potential future improvements to the current pipeline. We may first extract some characteristics of the task output, such as categorical criteria like \emph{problem\:type} and \emph{problem\:level}, and then potentially generate different criteria and quantify the task output differently based on these characteristics. 

\clearpage

\begin{table}[hb]
    \centering
        \caption{Comprehensive Verification Criteria for MathProblems.}

    \scalebox{0.5}{
    \begin{tabular}{>{\raggedright\arraybackslash}p{1.8cm}>{\raggedright\arraybackslash}p{5.8cm}>{\raggedright\arraybackslash}p{6.6cm}}
        \hline \hline
        \textbf{Criteria} & \textbf{Description} & \begin{center}
        \vspace{-1em}
            \textbf{Accepted Values}
                    \vspace{-1em}
        \end{center}  \\
        \hline
        efficiency & The conciseness of the solution and the use of the most efficient method to solve the problem.  & 
                              \vspace{-0.5em}
\begin{enumerate}[label=--,itemsep=0em,parsep=-0.2em,topsep=-4em]
        \item highly\_efficient (2)
        \item moderately\_efficient (1)
        \item inefficient (0)
        \end{enumerate} 
        \\[-1em]
        \hline
       accuracy & The correctness of the solution provided for the math problem. & 
                                       \vspace{-0.5em}
 \begin{enumerate}[label=--,itemsep=0em,parsep=-0.2em,topsep=-4em]
        \item 100\% - Completely correct (4)
        \item 75\% - Almost correct (3)
        \item 50\% - Mostly correct (2)
        \item 25\% - Partially correct (1)
        \item 0\% - Completely incorrect (0)
        \end{enumerate} 
       \\[-1em]
        \hline
               completeness & The extent to which the solution covers all aspects of the problem. & 
                                       \vspace{-0.5em}
 \begin{enumerate}[label=--,itemsep=0em,parsep=-0.2em,topsep=-4em]
            \item 100\% - Fully complete (4)
            \item 75\% - Almost complete (3)
            \item 50\% - Mostly complete (2)
            \item 25\% - Partially complete (1)
            \item 0\% - Not complete" (0)
        \end{enumerate} 
       \\[-1em]
        \hline
               clarity & The ease with which the solution can be understood by the target audience. & 
                                       \vspace{-0.5em}
 \begin{enumerate}[label=--,itemsep=0em,parsep=-0.2em,topsep=-4em]
            \item 100\% - Very clear (4)
            \item 75\% - Mostly clear (3)
            \item 50\% - Fairly clear (2)
            \item 25\% - Somewhat clear (1)
            \item 0\% - Not clear (0)
        \end{enumerate} 
       \\[-1em]
        \hline
               presentation & The organization and presentation of the solution, including proper use of notation, symbols, and formatting. & 
                                       \vspace{-0.5em}
 \begin{enumerate}[label=--,itemsep=0em,parsep=-0.2em,topsep=-4em]
            \item excellent (2)
            \item fair (1)
            \item poor (0)
        \end{enumerate} 
       \\[-1em]
        \hline
               steps delineation & How well the solution breaks down the problem-solving process into smaller, manageable steps. & 
                                       \vspace{-0.5em}
 \begin{enumerate}[label=--,itemsep=0em,parsep=-0.2em,topsep=-4em]
            \item 100\% - All steps delineated (4)
            \item 75\% - Most steps delineated (3)
            \item 50\% - Some steps delineated (2)
            \item 25\% - Few steps delineated (1)
            \item 0\% - No steps delineated (0)
        \end{enumerate} 
       \\[-1em]
        \hline
               response time & The time taken to provide the solution & 
                                       \vspace{-0.5em}
 \begin{enumerate}[label=--,itemsep=0em,parsep=-0.2em,topsep=-4em]
            \item >5 min (5) \hspace{2.7em} $-$ 3-5 min (4)
            \item 1-3 min (3) \hspace{2.4em} $-$ 31-60 sec (2)
            \item   16-30 sec (1) \hspace{1.8em} $-$ 0-15 sec (0)
        \end{enumerate} 
       \\[-1em]
        \hline
               notations & The notations used in the problem solution are appropriate and consistent. & 
                                       \vspace{-0.5em}
 \begin{enumerate}[label=--,itemsep=0em,parsep=-0.2em,topsep=-4em]
            \item consistent (2)
            \item mostly consistent (1)
            \item inconsistent (0)
        \end{enumerate} 
       \\[-1em]
        \hline
               steps explanation & The extent to which each step in the solution is explained. & 
                                       \vspace{-0.5em}
 \begin{enumerate}[label=--,itemsep=0em,parsep=-0.2em,topsep=-4em]
            \item all steps (4)
            \item most steps (3)
            \item half steps (2)
            \item some steps (1)
            \item none (0)
        \end{enumerate} 
       \\[-1em]
        \hline
               error handling & How well the system identifies and addresses possible errors in the problem & 
                                       \vspace{-0.5em}
 \begin{enumerate}[label=--,itemsep=0em,parsep=-0.2em,topsep=-4em]
            \item Handled all errors (4)
            \item Handled most errors (3)
            \item Handled some errors (2)
            \item Handled very few errors (1)
            \item Ignored all errors (0)
        \end{enumerate} 
       \\[-1em]
        \hline
               use of methods & The use of relevant techniques and concepts to address and solve the math problem. & 
                                       \vspace{-0.5em}
 \begin{enumerate}[label=--,itemsep=0em,parsep=-0.2em,topsep=-4em]
            \item Excellent use (2)
            \item Adequate use (1)
            \item Poor use (0)
        \end{enumerate} 
       \\[-1em]
        \hline
               level appropriateness & How well-suited the solution provided by the system is for the given problem's level & 
                                       \vspace{-0.5em}
 \begin{enumerate}[label=--,itemsep=0em,parsep=-0.2em,topsep=-4em]
            \item Highly appropriate (4)
            \item Appropriate (3)
            \item Moderately appropriate (2)
            \item Slightly appropriate (1)
            \item Not appropriate (0)
        \end{enumerate} 
       \\[-1em]
        \hline
               solution depth & The depth of the solution provided in terms of showing all steps and important calculations & 
                                       \vspace{-0.5em}
 \begin{enumerate}[label=--,itemsep=0em,parsep=-0.2em,topsep=-4em]
            \item Extremely detailed (3)
            \item Detailed (2)
            \item Moderate (1)
            \item Superficial (0)
        \end{enumerate} 
       \\[-1em]
        \hline
               terminology & Correct and consistent use of mathematical terminology in the explanations & 
                                       \vspace{-0.5em}
 \begin{enumerate}[label=--,itemsep=0em,parsep=-0.2em,topsep=-4em]
            \item Appropriate (2)
            \item Mostly appropriate (1)
            \item Inappropriate (0)
        \end{enumerate} 
       \\[-1em]
        \hline
               reliability & The dependability of the procedure/algorithm used in providing the solution & 
                                       \vspace{-0.5em}
 \begin{enumerate}[label=--,itemsep=0em,parsep=-0.2em,topsep=-4em]
            \item Distrusted (2)
            \item Mostly Trusted (1)
            \item Trusted (0)
        \end{enumerate} 
       \\[-1em]
        \hline
               calculation error & Presence of any computational or mathematical mistakes in the solution & 
                                       \vspace{-0.5em}
 \begin{enumerate}[label=--,itemsep=0em,parsep=-0.2em,topsep=-4em]
            \item No errors (2)
            \item Minor errors (1)
            \item Major errors (0)
        \end{enumerate} 
       \\[-1em]
        \hline
               creativity & Novel approach or method used in providing the solution & 
                                       \vspace{-0.5em}
 \begin{enumerate}[label=--,itemsep=0em,parsep=-0.2em,topsep=-4em]
            \item exceptionally novel (2)
            \item moderately novel (1)
            \item standard (0)
        \end{enumerate} 
       \\[-1em]
        \hline
               relevance & The solution should focus on solving the given problem and avoid irrelevant information or steps. & 
                                       \vspace{-0.5em}
 \begin{enumerate}[label=--,itemsep=0em,parsep=-0.2em,topsep=-4em]
            \item Highly relevant (2)
            \item Moderately Relevant (1)
            \item Irrelevant (0)
        \end{enumerate} 
       \\[-1em]
        \hline
               simplification & The degree to which the solution simplifies the problem while maintaining accuracy & 
                                       \vspace{-0.5em}
 \begin{enumerate}[label=--,itemsep=0em,parsep=-0.2em,topsep=-4em]
            \item Completely (3)
            \item Mostly (2)
            \item Partially (1)
            \item Not at all (0)
        \end{enumerate} 
       \\[-1em]
        \hline
               handling constraints & The accuracy of the solution in addressing given constraints & 
                                       \vspace{-0.5em}
 \begin{enumerate}[label=--,itemsep=0em,parsep=-0.2em,topsep=-4em]
            \item Fully respected (2)
            \item Partially respected (1)
            \item Disregarded (0)
        \end{enumerate} 
       \\[-1em]
        \hline
               problem type & The type of the math problem & 
                                       \vspace{-0.5em}
 \begin{enumerate}[label=--,itemsep=0em,parsep=-0.2em,topsep=-4em]
            \item Excellent (4) \hspace{2.2em} $-$Good (3)
            \item Average (2) \hspace{2.7em} $-$Poor (1)
            \item Terrible (0)
        \end{enumerate} 
       \\[-1em]
        \hline
        adaptability & Adaptability refers to the ability of the solution provided to be modified and adjusted to alternative or related problems. & 
                                       \vspace{-0.5em}
 \begin{enumerate}[label=--,itemsep=0em,parsep=-0.2em,topsep=-4em]
            \item Other (11) 
            \item Logic (10)
            \item Topology (9)
            \item Differential Equations (8)
            \item Linear Algebra (7)
            \item Number Theory (6)
            \item Combinatorics (5) \hspace{0em} $-$Statistics (4)
            \item Calculus (3) \hspace{2.3em} $-$Trigonometry (2)
            \item Geometry (1) \hspace{1.8em} $-$Algebra (0)

        \end{enumerate}
    \\[-1em]
        \hline
        problem level & The difficulty level of the math problem &              
                                       \vspace{-0.5em}
 \begin{enumerate}[label=--,itemsep=0em,parsep=-0.2em,topsep=-4em]
            \item Level 5 (4)  \hspace{2.7em} $-$Level 4 (3)
            \item Level 3 (2)   \hspace{2.7em} $-$Level 2 (1)
            \item Level 1 (0)
 
        \end{enumerate}
    \\[-1em]
        \hline
        solution approach & Appropriateness of the solution approach used & 
                                       \vspace{-0.5em}
 \begin{enumerate}[label=--,itemsep=0em,parsep=-0.1em,topsep=-4em]
            \item Appropriate (2)
            \item Questionable (1)
            \item Inappropriate (0)
        \end{enumerate}
        \\[-1em]
        \hline
        correct reasoning &The extent to which the systems response demonstrates correct mathematical reasoning. & 
        \vspace{-0.5em}
  \begin{enumerate}[label=--,itemsep=0em,parsep=-0.1em,topsep=-4em]
            \item 100\% (4)
            \item 75\% (3)
            \item 50\% (2)
            \item 25\% (1)
            \item 0\% (0)
        \end{enumerate}
        \\[-1em]
         \hline \hline
    \end{tabular}
    }
    \label{tab:full-criteria-math}
\end{table}

\begin{table}[htp]
    \centering
        \caption{Comprehensive Verification Criteria for ALFWorld Housholding
Tasks. }

    \scalebox{0.5}{
    \begin{tabular}{>{\raggedright\arraybackslash}p{2.1cm}>{\raggedright\arraybackslash}p{6.3cm}>{\raggedright\arraybackslash}p{5.7cm}}
        \hline \hline
        \textbf{Criteria} & \textbf{Description} & \begin{center}
        \vspace{-1em}
            \textbf{Accepted Values}
                    \vspace{-1em}
        \end{center}  \\
        \hline
        task completion & Degree to which the task is completed successfully  & 
                              \vspace{-0.5em}
\begin{enumerate}[label=--,itemsep=0em,parsep=-0.2em,topsep=-4em]
        \item 100\% (4)
        \item 75\% (3)
        \item 50\% (2)
        \item 25\% (1)
        \item 0\% (0)
        \end{enumerate} 
        \\[-1em]
        \hline
       action validity & Actions must be chosen from the given available actions, with illegitimate actions taken into account & 
                                       \vspace{-0.5em}
 \begin{enumerate}[label=--,itemsep=0em,parsep=-0.2em,topsep=-4em]
         \item all\_legal (3)
         \item one\_illegal (2)
         \item two\_illegal (1)
        \item three\_or\_more\_illegal (0)
        \end{enumerate} 
       \\[-1em]
        \hline
               thought process & The quality of the thought process and planning throughout the task & 
                                       \vspace{-0.5em}
 \begin{enumerate}[label=--,itemsep=0em,parsep=-0.2em,topsep=-4em]
            \item excellent (3)
            \item good (2)
            \item fair (1)
            \item poor(0)

        \end{enumerate} 
       \\[-1em]
        \hline
               systematic search & How systematically the player searched for items and target locations & 
                                       \vspace{-0.5em}
 \begin{enumerate}[label=--,itemsep=0em,parsep=-0.2em,topsep=-4em]
            \item excellent (3)
            \item good (2)
            \item moderate (1)
            \item poor (0)
        \end{enumerate} 
       \\[-1em]
        \hline
               interaction flow & The smoothness and continuity of interactions with the environment & 
                                       \vspace{-0.5em}
 \begin{enumerate}[label=--,itemsep=0em,parsep=-0.2em,topsep=-4em]
            \item smooth (2)
            \item some\_disruptions (1)
            \item frequent\_disruptions (0)
        \end{enumerate} 
       \\[-1em]
        \hline
               task time & The time taken to accomplish the task & 
                                       \vspace{-0.5em}
 \begin{enumerate}[label=--,itemsep=0em,parsep=-0.2em,topsep=-4em]
            \item very\_fast (3)
            \item fast (2)
            \item average (1)
            \item slow (0)
        \end{enumerate} 
       \\[-1em]
        \hline
               planning strategy & Quality of the devised plan for completing the task & 
                                       \vspace{-0.5em}
 \begin{enumerate}[label=--,itemsep=0em,parsep=-0.2em,topsep=-4em]
            \item excellent (3)
            \item good (2)
            \item fair (1)
            \item poor (0)
        \end{enumerate} 
       \\[-1em]
        \hline
               action efficiency & Efficiency of the chosen actions in solving the task & 
                                       \vspace{-0.5em}
 \begin{enumerate}[label=--,itemsep=0em,parsep=-0.2em,topsep=-4em]
            \item very high (3)
            \item high (2)
            \item moderate (1)
            \item low (0)
        \end{enumerate} 
       \\[-1em]
        \hline
               response format & Adherence to the required response format & 
                                       \vspace{-0.5em}
 \begin{enumerate}[label=--,itemsep=0em,parsep=-0.2em,topsep=-4em]
            \item correct (2)
            \item partially correct (1)
            \item incorrect (0)
        \end{enumerate} 
       \\[-1em]
        \hline
               adaptability to feedback & Ability to adapt and modify the plan based on the environment's feedback & 
                                       \vspace{-0.5em}
 \begin{enumerate}[label=--,itemsep=0em,parsep=-0.2em,topsep=-4em]
            \item very high (3)
            \item high (2)
            \item moderate (1)
            \item low (0)
        \end{enumerate} 
       \\[-1em]
        \hline
               termination judgement & Proper judgment of when to reply with 'TERMINATE' & 
                                       \vspace{-0.5em}
 \begin{enumerate}[label=--,itemsep=0em,parsep=-0.2em,topsep=-4em]
            \item correct (2)
            \item partially correct (1)
            \item incorrect (0)
        \end{enumerate} 
       \\[-1em]
        \hline
               efficiency & Assesses the number of steps taken in comparison to the minimum possible steps required to complete the task & 
                                       \vspace{-0.5em}
 \begin{enumerate}[label=--,itemsep=0em,parsep=-0.2em,topsep=-4em]
            \item optimal (3)
            \item near\_optimal (2)
            \item average (1)
            \item below\_average (0)
        \end{enumerate} 
       \\[-1em]
        \hline
               problem solving & The ability to quickly identify and adapt to changes in the environment during task execution & 
                                       \vspace{-0.5em}
 \begin{enumerate}[label=--,itemsep=0em,parsep=-0.2em,topsep=-4em]
            \item fast\_adaptation (3)
            \item moderate\_adaptation (2)
            \item slow\_adaptation (1)
            \item no\_adaptation (0)
        \end{enumerate} 
       \\[-1em]
        \hline
               target handling & How well the player followed instructions for handling the target household & 
                                       \vspace{-0.5em}
 \begin{enumerate}[label=--,itemsep=0em,parsep=-0.2em,topsep=-4em]
            \item excellent (3)
            \item good (2)
            \item moderate (1)
            \item poor (0)
        \end{enumerate} 
       \\[-1em]
        \hline
               environment understanding & The ability to understand the provided environment description and identify relevant objects & 
                                       \vspace{-0.5em}
 \begin{enumerate}[label=--,itemsep=0em,parsep=-0.2em,topsep=-4em]
            \item excellent (3)
            \item good (2)
            \item fair (1)
            \item poor (0)
        \end{enumerate} 
       \\[-1em]
        \hline
               compliance with instructions & Adherence to specific rules and instructions such as reply formatting and termination conditions & 
                                       \vspace{-0.5em}
 \begin{enumerate}[label=--,itemsep=0em,parsep=-0.2em,topsep=-4em]
            \item compliant (2)
            \item partially compliant (1)
            \item non-compliant (0)
        \end{enumerate} 
       \\[-1em]
        \hline
               legal actions & Selecting actions from the given available actions and avoiding illegal actions & 
                                       \vspace{-0.5em}
 \begin{enumerate}[label=--,itemsep=0em,parsep=-0.2em,topsep=-4em]
            \item excellent (4)
            \item good (3)
            \item average (2)
            \item below\_average (1)
            \item poor (0)
        \end{enumerate} 
       \\[-1em]
        \hline
               target acquisition & Acquiring the target household object immediately after finding it & 
                                       \vspace{-0.5em}
 \begin{enumerate}[label=--,itemsep=0em,parsep=-0.2em,topsep=-4em]
            \item excellent (3)
            \item good (2)
            \item fair (1)
            \item poor (0)
        \end{enumerate} 
       \\[-1em]
        \hline
               format adherence & The extent to which the output format is strictly followed & 
                                       \vspace{-0.5em}
 \begin{enumerate}[label=--,itemsep=0em,parsep=-0.2em,topsep=-4em]
            \item Correct format (2)
            \item Minor format issues (1)
            \item Incorrect format (0)
        \end{enumerate} 
       \\[-1em]
        \hline
               problem understanding & Understanding of the given task and relevance of the environment & 
                                       \vspace{-0.5em}
 \begin{enumerate}[label=--,itemsep=0em,parsep=-0.2em,topsep=-4em]
            \item 3 - Fully understood (3)
            \item 2 - Adequately understood (2)
            \item 1 - Partially understood (1)
            \item 0 - Not understood (0)
        \end{enumerate} 
       \\[-1em]
        \hline
               action selection & Choosing the appropriate sequence and type of actions & 
                                       \vspace{-0.5em}
 \begin{enumerate}[label=--,itemsep=0em,parsep=-0.2em,topsep=-4em]
            \item 3 - Optimal selection (3)
            \item 2 - Good selection (2)
            \item 1 - Somewhat acceptable selection (1)
            \item 0 - Poor selection (0)
        \end{enumerate} 
        \\[-1em]
         \hline \hline
    \end{tabular}
    }
    \label{tab:full-criteria-alf}
\end{table}

\clearpage

\end{document}